\documentclass[conference]{IEEEtran}
\IEEEoverridecommandlockouts
\usepackage{cite}
\usepackage{amsmath,amssymb,amsfonts}
\usepackage{algorithmic}
\usepackage{graphicx}
\usepackage{textcomp}
\usepackage{xcolor}

\usepackage{todonotes}
\usepackage{subcaption}
\usepackage{url}
\usepackage{hyperref}
\usepackage{gensymb}
\usepackage{multirow}
\usepackage[pdf]{graphviz}
\usepackage[normalem]{ulem} 

\usepackage{booktabs}
\usepackage{threeparttable}
\usepackage{multirow}
\usepackage{siunitx}
\usepackage{pgfplots}
\pgfplotsset{compat=1.18}
\def\BibTeX{{\rm B\kern-.05em{\sc i\kern-.025em b}\kern-.08em
		T\kern-.1667em\lower.7ex\hbox{E}\kern-.125emX}}

\begin{document}
	
	\title{Have I Solved This Before? Retrieving Similar Segmentation Problems for Evolutionary Learning}
	
	\author{
		\IEEEauthorblockN{
			\IEEEauthorrefmark{1}Andreas Margraf,
			\IEEEauthorrefmark{2}Henning Cui
			and
			\IEEEauthorrefmark{2}J\"org H\"ahner
		}
		\IEEEauthorblockA{
			\IEEEauthorrefmark{1}
			\IEEEauthorrefmark{2}
			\textit{Institute of Computer Science},
			\textit{University of Augsburg, Germany}\\ 
			Email: \IEEEauthorrefmark{1}andreas.margraf@gmail.com,
			\IEEEauthorrefmark{2}\{henning.cui,joerg.haehner\}@uni-a.de}
	}
	
	\maketitle
	
	\thispagestyle{plain}
	\pagestyle{plain}
	
	\begin{abstract}
		Reliable integration and solid configuration of monitoring systems constitute a fundamental prerequisites for achieving high efficiency and productivity in contemporary manufacturing environments.
		Design decisions on sensor type and system architecture have to be made at an early stage and under comparably high uncertainty. This work investigates a research direction that deviates from the traditional monitoring-system development process by shifting the attention from algorithm design to a deeper analysis of the inspection problem. In contrast to traditional design cycles, this paper proposes to gradually collect knowledge and store it in an abstract system model. This enables the retrieval of similar solutions for future use cases, preventing the need for expensive model training from scratch and allowing instead for the incremental refinement of existing base configurations. Reuse of previously generated pipelines reduces the risk of late and costly revisions. As there is little knowledge on cross-domain transferability of filter pipelines, this study analyzes the potential of retrieving filter pipelines to transfer them to different but similar segmentation problems. Finally, we statistically analyze the benefits of this `transfer learning' variant which is predominantly applied to image segmentation problems. In addition, we discuss how simple models help balancing the trade-off between complexity, technical requirements, and reliability in the design process.
	\end{abstract}
	
	
	\section{Introduction}\label{sec:introduction}
	Quality assurance and process monitoring are considered a key factor in production management and have attracted more attention as machine parks and equipment share machine data at an ever-larger scale. Concepts such as edge computing or Industrial Internet of Things (IIoT) strive for at least one common goal: increased interconnection between components, devices, and sensors as part of large networks. A primary objective is to enable systems to develop self-adaptive, self-organizing, or self-configuring properties. Evolutionary Algorithms (EAs) have long served as a method to mimic this behavior: They are known as flexible, nature-inspired heuristics that can be applied to optimization, modeling, and simulation problems.
	The detection of defects such as structural damage, structural misalignments, pattern anomalies, shape deviations, or missing units incorporate some of the most common faults in industrial production which have been pointed out in respective studies ~\cite{Geinitz2016,Margraf2017a}. This research work has shown that Cartesian Genetic Programming (CGP) in particular allows to create programs that process industrial images efficiently in the edge computing sense.
	
	\subsection{Solving Segmentation Tasks using Efficient Algorithms}
	The spread of large foundation models has unlocked new opportunities across diverse domains, from computer vision to language processing ~\cite{Halevy2009,Hestness2017,Jain2020,Sevilla2022,Goldblum2023}. In contrast to this tendency towards `one size fits all' design principles, we propose to break loose from this single-sided thinking and focus on the design process to improve the selection and configuration of small-size models. In order to target highly specialized applications that require more resource efficiency than large models can provide, e.g. ASICS, FPGA, or low-power, low-latency systems hardened for tough conditions. 
	We consider it noteworthy to promote the perspective of systems that improve and adapt over time in an autonomous manner following the spirit of Organic Computing (OC) and Autonomous Computing (AC). However, this is only one side of the medal: This strategy applies to environments with uncertain conditions `in the wild'. In an industrial context, though, uncertainties that arise from the immediate surroundings are sought to be minimized:
	
	Instead of redesigning a manufacturing line, which raises operating cost, a self-configuring system may handle changes of external parameters with ease, if equipped with an on-demand optimization component, as has been proposed by Stein et al. \cite{Stein2018}. Furthermore, this study is widely based on filter pipeline evolution as it has been proposed in recent work by Margraf et al. \cite{Margraf2023a,Margraf2025}. This study is set out to examine the following research questions (RQ):
	\begin{itemize}
		\item \textbf{RQ1:} Which method allows to retrieve the best filter pipeline given the process data of an industrial monitoring task?
		\item \textbf{RQ2:} How can solutions from previous applications be retrieved and applied to new problems based on similarity?
	\end{itemize}
	
	\subsection{Self-adaptivity in Modular Units}
	In contrast to Mueller-Schloer's \textit{Organic Computing (OC)} \cite{muller2011organic} or Kephart's \textit{Autonomic Computing (AC)} \cite{Kephart2003} which are strategies to handle the potentially unpredictably behavior of large systems, recent research proposes large models trained on a myriad of data to manage the complexity of modern information flows \cite{Kephart2003}.
	
	This study borrows the principles of OC and AC and applies a new strategy we denote `retrieve and adapt' to modular units with limited complexity and configuration space. We let existing and realistic industrial process monitoring (PM) systems learn appropriate configurations on small computing units for which a high ability to adapt is required.
	
	In this paper, we introduce a methodology for adaptive design and evolution that applies to industrial monitoring systems. Furthermore, we propose a strategy that uses model and algorithm selection to allow developers to create systems that are substantially more efficient and use fewer resources even when complex patterns are analyzed in the data. For this purpose, we propose \textit{a) keeping model configurations and filter pipelines} in a database for any previously solved task and \textit{b) retrieving similar configurations} and applying them to new, unseen problems to save a considerable amount of design time in system development. Finally, we evaluate our hypotheses that similar pipelines perform equally on similar tasks in an experimental analysis.
	\section{Structure}
	The remainder of this paper is organized as follows:
	Section \ref{sec:introduction} begins by outlining the challenge of monitoring system design and configuration. After summing up the related work in section \ref{sec:relatedwork}, the following section \ref{sec:approach} presents the methodology employed.
	An in-depth analysis of the pipeline retrieval approach is presented in Section \ref{sec:performance-based}. In addition, Section \ref{sec:discussion} offers explanations for our findings and critically reflects on the pipeline retrieval approach. Finally, section \ref{sec:conclusion} recapitulates the results as well as the experimental analysis before giving an outlook on future work.
	
	\section{Related Work}\label{sec:relatedwork}
	Several publications address research topics closely related to this work and are discussed in the following section:
	Early work in this area focused on evaluating the efficiency of deep learning image classification as proposed by Hernandez et al. \cite{Hernandez2020}. Ke et al. \cite{Ke2021} investigated multitask deep learning while Pfisterer et al. \cite{Pfisterer2018} suggested learning default values from prior results. 
	
	Further research proposed genetic programming for the segmentation of data streams: Thus, CGP was applied to image filters \cite{Harding2013} and signal classification \cite{Kaufmann2013}. 
	Also, Cui et al. proposed reorder strategies for CGP to mitigate positional bias \cite{Cui2023, Cui2024a}. Interestingly, further examinations show that CGP proves robust on noisy datasets \cite{Cui2024b}.
	
	A first step towards image similarity was taken using mutual information of regions as proposed by Russakoff et al. \cite{Russakoff2004} and extended by modeling complexity with component analysis for image retrieval \cite{Perkioe2009}, difficulty scores \cite{Ionescu2016}, spatial information \cite{Yu2013} and complexity by independent component analysis \cite{Perkioe2009}. Interestingly, Redies et al. \cite{Redies2012} followed a slightly different path, proposing measures to compute the aesthetics of color photographs. Not to forget, Domingos et al. \cite{Domingos2012} offered an overview on key machine learning practices and metrics for image segmentation of various complexity. It should also be mentioned that the Histogram of Oriented Gradients proved useful for image similarity computation \cite{Dalal2005}.
	
	In an effort to compare images based on their complexity,, Yu et al. proposed to consider spatial information as a metric \cite{Yu2013}. Ionescu et al. examined human ability to determine difficulty of image segmentation tasks \cite{Ionescu2016}.
	While Ivanovic et al. \cite{Ivanovici2020} examined the correlation between complexity measures and the number of an image's segment. Furthermore, Margraf et al. explored the similarity between image complexity and CGP fitness on industrial image datasets \cite{Margraf2023a} and introduced an approach that allows CGP to improve the resource efficiency of industrial signal and image processing \cite{Margraf2025}. Furthermore, this study contributes to the research work recently conducted by Margraf \cite{Margraf2025}. 
	Pitfalls in the field of algorithm configuration have been intensively explored by Eggensperger et al. \cite{Eggensperger2017}, while Kerschke et al. presented a review on automated algorithm selection \cite{Kerschke2018}. 
	
	Despite the number of publications in this field, the authors could not identify studies that discuss the combined analysis of complexity and similarity metrics for the retrieval of filter pipelines to be applied on monitoring solutions.
	
	\section{Pipeline Retrieval by Dataset Similarity}\label{sec:approach}
	For any high-level analysis of continuous data, it is essential to extract meaningful characteristics of the data. We distinguish between complexity, difficulty and similarity, noting they are mutually dependent. In this study, the term `data' is used as a general description for time series, signals and images. We distinguish between three key metrics:
	\begin{itemize}
		\item \textit{Difficulty:} Required effort to perform a given task, e.g. segmentation of certain objects in an image
		\item \textit{Complexity:} Structure, variability or information content of an image or data frame
		\item \textit{Similarity:} Degree to which two or several images or data frames share common or equal characteristics
	\end{itemize}
	In this study, the aforementioned metrics are employed to analyze the dataset within the context in which they were originally developed. The notion \textit{complexity} characterizes the structural information present in an image and provides a means to better understand its content. In addition, \textit{difficulty} is inherently task-dependent. It must therefore be considered separately as it only partially correlates with complexity. For instance, background and foreground separation is typically far less demanding than extracting fine fiber structures from an image. Finally, \textit{similarity} points to a completely different aspect of this study: As we assume that a given algorithm or model has been applied to previous datasets with different tasks, we want to analyze how similar a historic task is in comparison to a new problem. We hypothesize that algorithms that were successfully applied in the past, particularly in image processing applications, yield results of comparable quality when applied to a new, but similar task.
	
	\subsection{Three-Tier Approach for Pipeline Transferability}
	
	This research aims to address the following key question: 
	
	\textit{How well does a filter pipeline, once generated, perform on a different but similar segmentation problem?} 
	
	To explore this, datasets from different industrial use cases have been identified (\textit{cf. Appdx. Tab. \ref{tab:datasets}}) that bear typical surface defects on different material. In order to evaluate the potential of the proposed filter and pipeline-based image processing, the following three-tier methodology is applied:
	
	\begin{enumerate}
		\item \textit{Similarity computation:} We compute pairwise similarity between the 38 datasets based on CNN embeddings (ResNet), 
		and complementary measures 
		(entropy, texture, edge density, frequency, superpixels).
		\item \textit{Cross-application of pipelines:} We evaluate cross-dataset performance by applying CGP-generated pipelines trained on a source dataset 
		to all other datasets, resulting in a matrix of MCC scores. 
		From this, we derive aggregated measures for mean/best MCC and pipeline-dataset many-to-many transferability.
		\item \textit{Statistical performance analysis:} 
		We analyze cross-application performance by correlation, 
		including bin-wise evaluation by similarity level. 
		Furthermore, we examine the similarity-performance relationship by means of ordinary least squares (OLS) and logistics regression, as well as statistical significance measures.	
	\end{enumerate}
	
	This allows both a systematic exploration of pipeline transferability and quantification of its performance across industrial domains. For a better understanding of the dataset context and pipeline composition, the following section will give a short introduction on how CGP is applied to create filter pipelines for image processing.
	
	\subsection{Effects of Complexity and Difficulty on Algorithm Design}
\begin{figure}[t!]
		\centering
		\includegraphics[width=1.0\columnwidth]{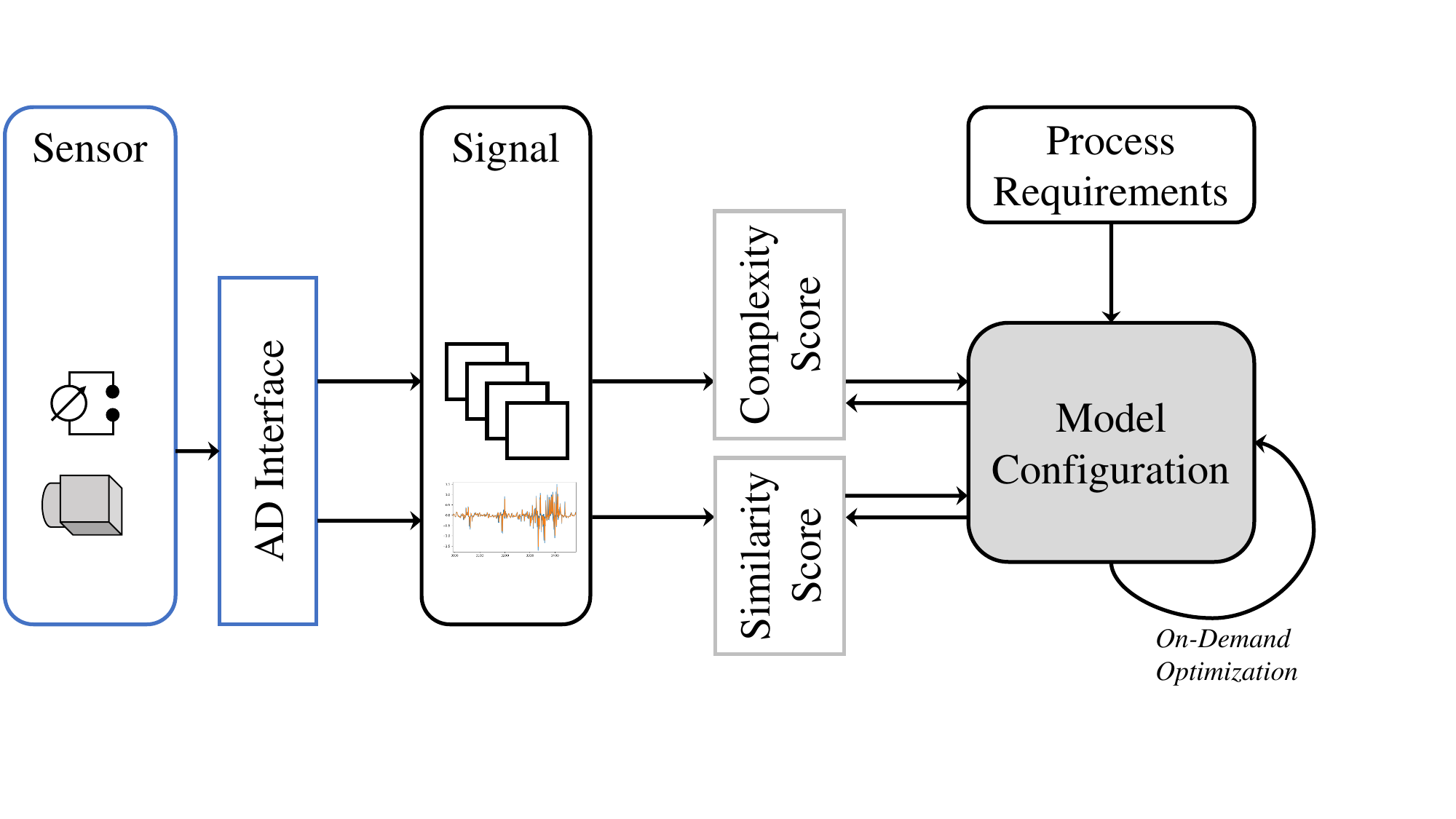}
		\caption{Envisioned concept: Algorithms and models are selected not only according to input data, but also by sensor type, sensor configuration and the expected, application-specific output; model reconfiguration is executed on demand, i.e. for changes of process requirements or variations of the inspection object's properties.}\label{fig:sensor_model_config}
	\end{figure}

	The first part of the methodology presented in this study builds on the principle of pipeline learning and reuse (cf. Fig. \ref{fig:sensor_model_config}). On the one hand, it relies on selecting the most suitable model for a given segmentation task (\textit{algorithm selection}). However, the effectiveness of training strongly depends on model complexity; therefore, this choice must be guided by the complexity of the task—or of the underlying dataset in the first place. From a data-analysis perspective, complexity acts as an indicator of the degree of randomness or regularity in the data, as well as of the amount of non-linear information conveyed by a stream of data frames per unit time.
	
	The second part is based on the concept of \textit{transfer learning}. This is achieved by applying a parameterized pipeline, learned from previous tasks, to a new but similar problem. The retrieval of suitable pipelines based on task similarity is referred to as \textit{semantic search} in the computer vision context. Such similarity-based retrieval mechanisms are widely used in recommendation systems, where results are returned according to user preferences or contextual similarity. Within the methodology we propose in this study, transfer learning is employed to improve the efficiency of pipeline evolution by reusing previously learned and semantically related solutions.
	
	Previous work by Margraf et al. investigated the complexity of images and their annotated regions in order to determine a priori whether small, computationally efficient filter cascades are sufficient to solve a given segmentation problem \cite{Margraf2023a, Margraf2025}. 
	
	Out of ten complexity measures, we identified the most promising: Histogram Entropy (HE), Texture Composition (TC), Edge Density (ED), Number of Superpixels (SP) and Fourier Frequency (FF). The definitions for the complexity measures can be found in Appdx. \ref{appx:image-complexity}. This is based on the publication by Margraf et al.~\cite{Margraf2023a,Margraf2025}. In this study, we add CNN image embeddings to the list. For this purpose, feature vectors are obtained by averaging CNN-based image embeddings, extracted by means of a ResNet-50 model (pretrained on ImageNet) across a small, representative subset of images, which is labeled 'CNN' in the following sections. 
	
	Earlier work by Margraf et al. compares  U-Net, SVM, and Random Forest models and reveals that data complexity serves as an indicator of comparable model performance \cite{Margraf2025}. This methodology enables the identification of use cases in which compact, automatically generated programs can achieve performance equal —or even superior to—deep learning models such as CNNs, RNNs, or transformer-based architectures.
	
	The work by Margraf et al. inspired the core design principle of the approach which we propose in this paper: Pipeline generation and parameterization is continuously improved as the database grows. 
	The retrieval is formulated as a similarity search based on a distance measure that quantifies the resemblance between a given inspection task and previously encountered problems (cf. Fig. \ref{fig:sensor_model_config}).
	
	The metrics discussed in this section provide a formal representation of the structural complexity of an image frame and the amount of information it potentially carries. However, it should be noted that the correlation between measured complexity and the perceived \textit{difficulty} of a segmentation task may differ in certain cases. Accordingly, our methodology is based solely on the descriptive nature of complexity and does not assume a direct equivalence to task difficulty.
	
	\subsection{Cosine Similarity and Cross-Application}
	In order to estimate the \textit{relatedness} between different image datasets, we compute cosine similarity for each complexity metric (CNN, HE, TC, ED, SP, FF) and learned feature embeddings of each dataset. 
	
	Let $A$ and $B$ denote the feature vectors of two datasets and let their cosine similarity be defined as:
	\begin{equation}
		Cos_{Sim}(A,B) = \frac{A \cdot B}{\|A\|\|B\|}    
	\end{equation}
	
	$Cos_{Sim}(A,B)$ describes the angular similarity between two sets in a high-dimensional feature space. A similarity score close to $1.0$ indicates that the datasets likely exhibit comparable structural or visual characteristics, while lower scores imply dissimilarity or none at all ($0.0$).
	Using this method, we compute pairwise cosine similarities between all $n=38$ sets.
	
	For each of the $38$ datasets, a segmentation pipeline is trained individually and then cross-applied to the remaining 37. This adds up to a total of $38 \times 37 = 1406$ transfer observations, excluding self-transfers.
	The predictive performance is quantified using the Matthews Correlation Coefficient (MCC), a robust metric for binary segmentation.
	Our hypothesis is that pipelines trained on datasets with high similarity to a target dataset will yield higher MCC values, compared to when applied to \textit{dissimilar} datasets.
	The set size of 1406 cross-dataset evaluations allows for a statistically meaningful analysis.
	
	\subsection{Statistical Analysis}
	As a first indicator, we compute the Spearman rank $r$ and Pearson $\rho$ correlation between cosine similarity and MCC values across all 1406 pairs to identify details of the similarity scores. This reveals in how far higher similarity is monotonically associated with higher performance. 
	A strong positive correlation coefficient ($r$, $\rho$) indicates a robust association between dataset similarity and transferability. Its statistical significance will be assessed using the p-values.
	For better understanding of different levels of correlation, the values are separated by coefficient bin partitions. The bins are split by a step size of $0.2$ as follows:
	\begin{equation}
		Cos_{Sim} \rightarrow bins ([-1.0, 0.3),[0.3,0.5), [0.5,0.7),[0.7,1.0])
	\end{equation}
	
	To quantify how consistently a pipeline transfers across datasets, we compute a transfer rate $TR_{i}$. It is defined per pipeline $i$ as the proportion of successful cross-applications, defined as the number of cross-application pipelines whose $MCC_{ij}$ value exceeds a threshold of $0.1$, over the total number of target datasets $N$:
	\begin{equation}
		TR_{i} = \frac{\# \{j:MCC_{ij} > 0.1 \}}{N}
	\end{equation}
	We then compute a robustness-adjusted transfer score $TS_i$ that equally considers the mean cross-application performance $\overline{MCC}_i$ with the transfer rate and penalizes unstable transfer behavior using the standard deviation $\sigma_i$:
	\begin{equation}
		TS_{i} = \overline{MCC_{i}} \times TR_{i} - \lambda \sigma_{i}
	\end{equation}
	with $\lambda = 0.25$. The value of $\lambda$ was chosen heuristically to penalize unstable transfer behavior while preserving the dominant contribution of average transfer quality. $TS_i$ therefore evenly represents the quality and robustness of transfer.
	
	We then fit a multivariate ordinary least squares (OLS) regression model in order to examine the influence of multiple similarity metrics on cross-dataset transferability:
	\begin{equation}
		MCC_{ij} = \beta_{0} + \sum_{k=1}^{p}{\beta_{k}x_{ij,k}} + \epsilon_{ij}
	\end{equation}
	where $MCC_{ij}$ denotes the cross-application performance obtained by applying a pipeline trained on dataset $i$ to dataset $j$, $x_{ij,k}$ represents the $k$-th similarity metric for the dataset pair, $\beta_k$ are regression coefficients, and $\epsilon_{ij}$ denotes the residual error term.
	The model reveals how strongly each similarity metric contributes to transfer performance while considering the influence of the remaining predictors. Statistical significance is exposed by the coefficient p-values and overall model fit statistics ($R^{2}$, adjusted $R^{2}$, AIC, and BIC).
	
	To model the probability of successful transfer, we define a binary target value $y$ indicating whether a pipeline achieves acceptable performance:
	\begin{equation}
		y = 
		\begin{cases}
			1, & \text{MCC} \geq 0.05 \\
			0, & \text{otherwise}
		\end{cases}
	\end{equation}
	
	A transfer is considered successful if a transfer pipeline $MCC$ exceeds $0.05$ on the target dataset as we consider lower values as an indication for no transferability.
	The probability of successful transfer is then modeled as:
	\begin{equation}
		P(y=1) = \frac{1}{1 + e^{-(\beta_{0} + \sum_{k}{\beta_{k}x_{k}})}}
	\end{equation}
	The logistic function transforms the linear predictor into a probability $P \in [0,1]$, thereby estimating which features increase the likelihood of successful cross-dataset generalization.
	
	The p-values of correlation coefficients and regression parameters point out whether the observed relationships are unlikely to occur by chance. As previously, the model quality is evaluated using the parameter $R^{2}$, and residual diagnostics.
	
	\section{Similarity-Based Performance Analysis}\label{sec:performance-based}
	For the performance analysis of the proposed pipeline retrieval approach, image recordings from different monitoring scenarios are used to evaluate the similarity algorithm.
	Although the 38 datasets cover a wide field and provide a large database, they are still sufficiently small to allow the computation of CNN-based embeddings and related complexity metrics.
	When scaling, computational cost becomes critical. This favors approaches like HoG or similar descriptors that offer a good balance between accuracy and efficiency. For further reading, our data and code is available on GitHub\footnote{cf. \url{https://tinyurl.com/estinspect}}.
	
	\begin{figure}[h]
		\centering
		\begin{subfigure}[h]{0.23\columnwidth}
			\includegraphics[width=\textwidth]{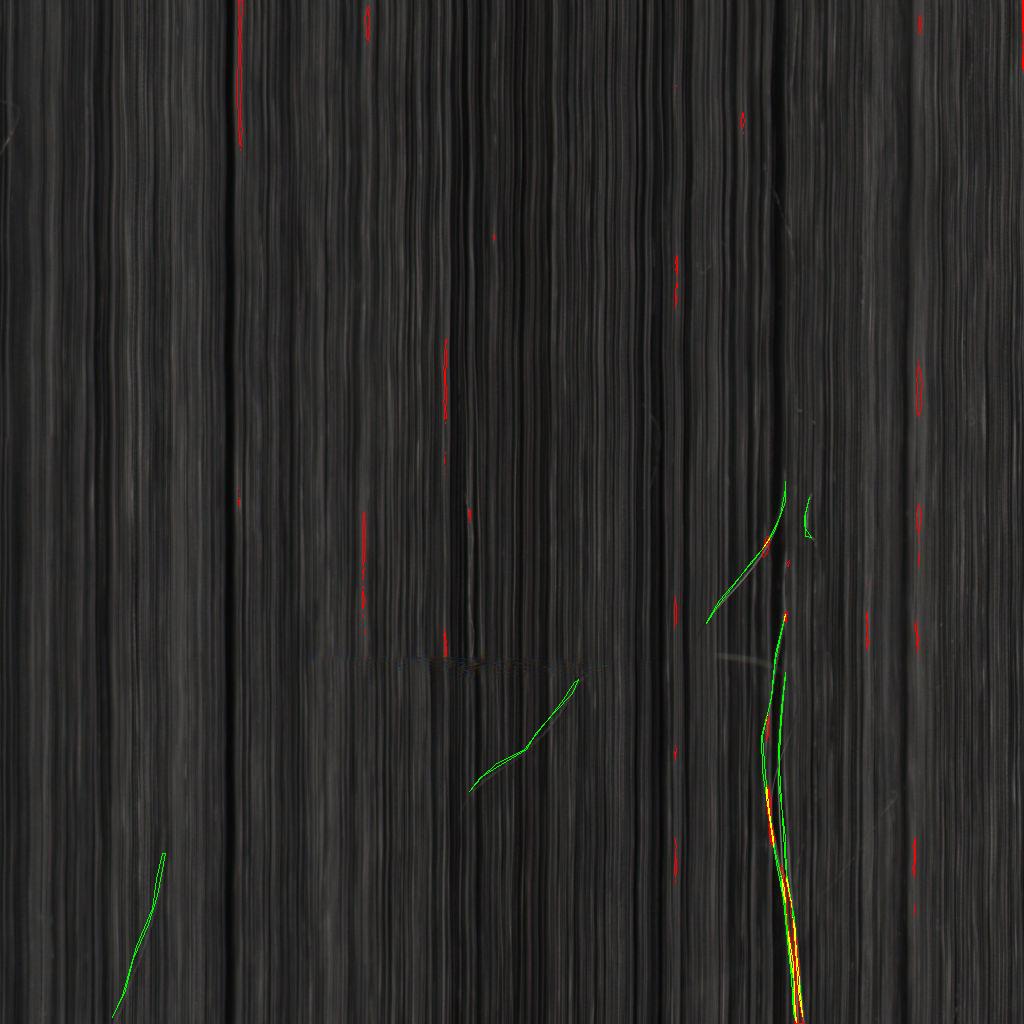}
		\end{subfigure}%
		\begin{subfigure}[h]{0.23\columnwidth}
			\includegraphics[width=\textwidth]{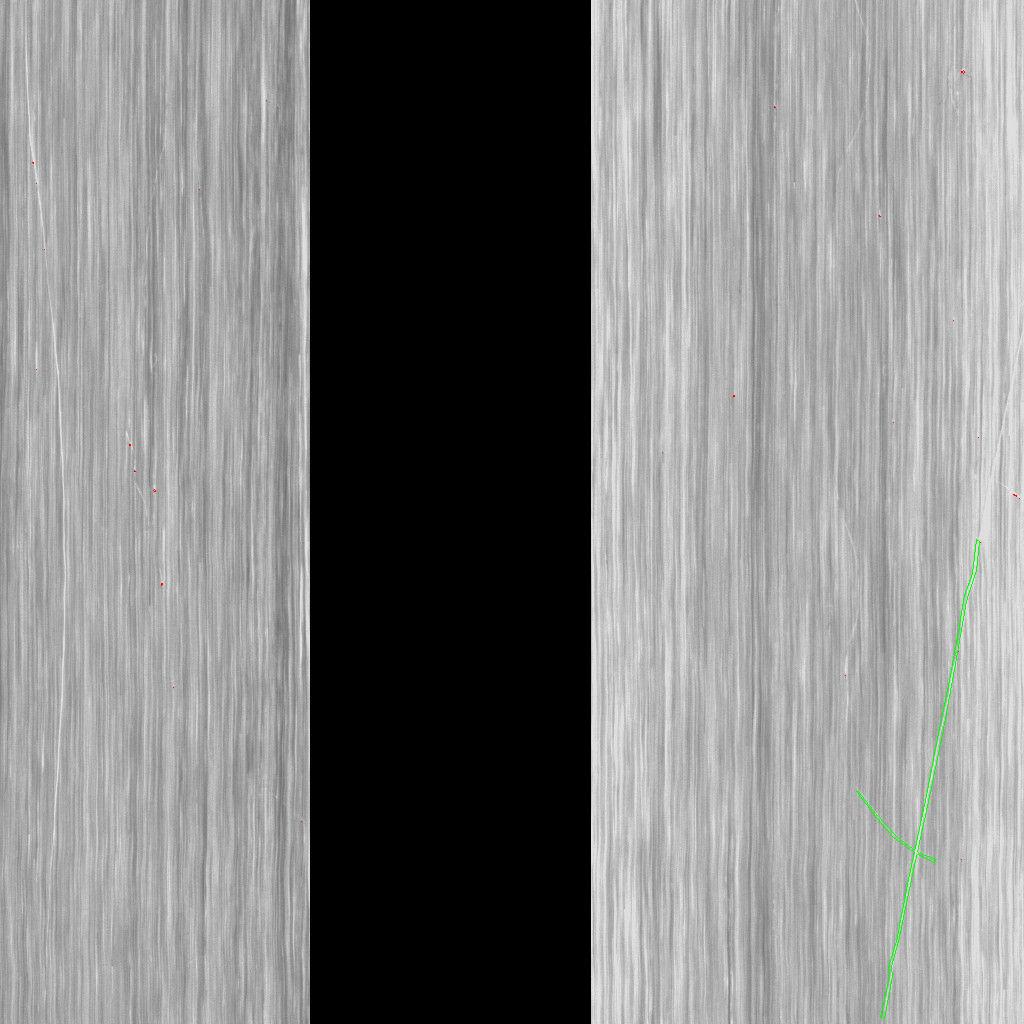}
		\end{subfigure}%
		\begin{subfigure}[h]{0.23\columnwidth}
			\includegraphics[width=\textwidth]{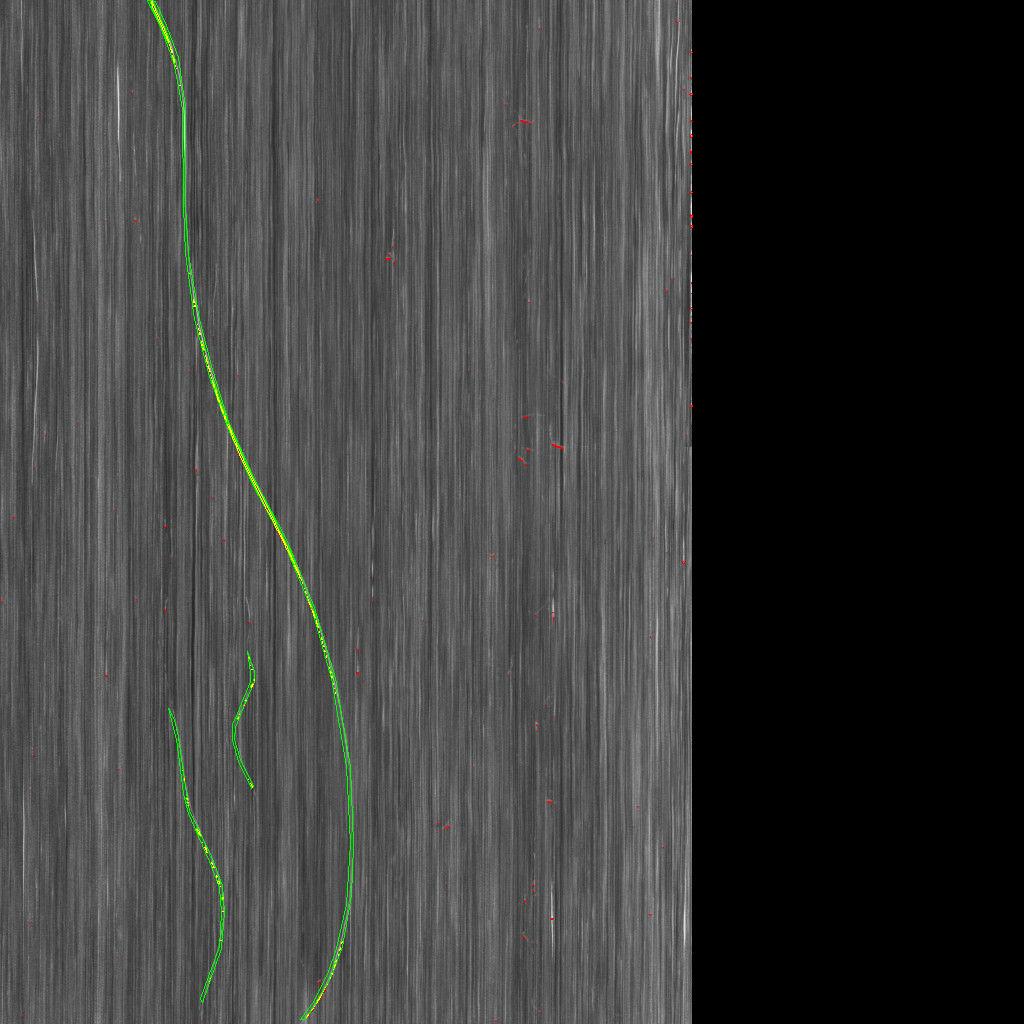}
		\end{subfigure}%
		\begin{subfigure}[h]{0.23\columnwidth}
			\includegraphics[width=\textwidth]{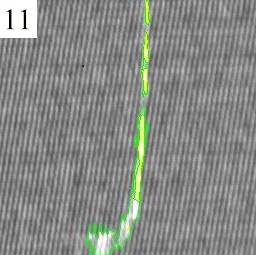}
		\end{subfigure}
		~
		\begin{subfigure}[h]{0.23\columnwidth}
			\includegraphics[width=\textwidth]{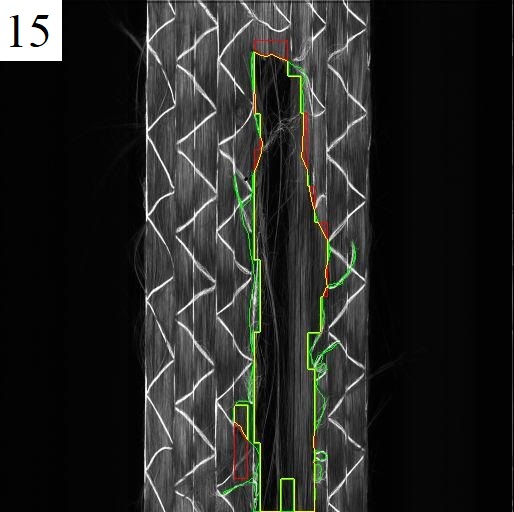}
		\end{subfigure}%
		\begin{subfigure}[h]{0.23\columnwidth}
			\includegraphics[width=\textwidth]{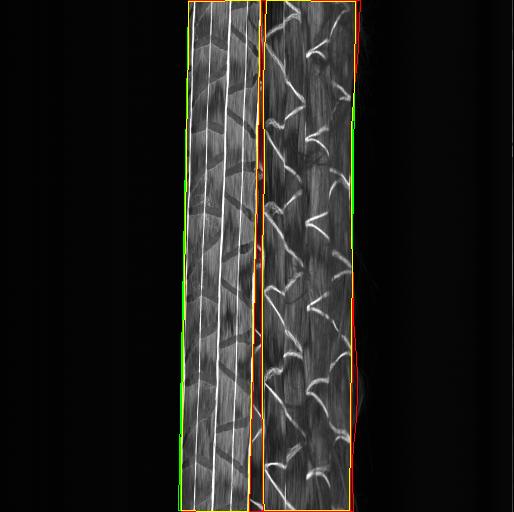}
		\end{subfigure}%
		\begin{subfigure}[h]{0.23\columnwidth}
			\includegraphics[width=\textwidth]{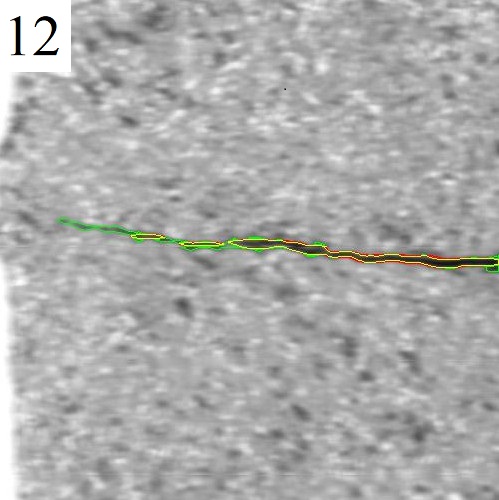}
		\end{subfigure}%
		\begin{subfigure}[h]{0.23\columnwidth}	
			\includegraphics[width=\textwidth]{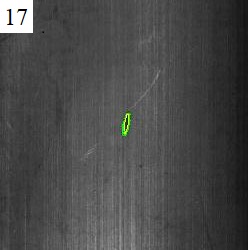}
		\end{subfigure}
		~
		\begin{subfigure}[h]{0.23\columnwidth}
			\includegraphics[width=\textwidth]{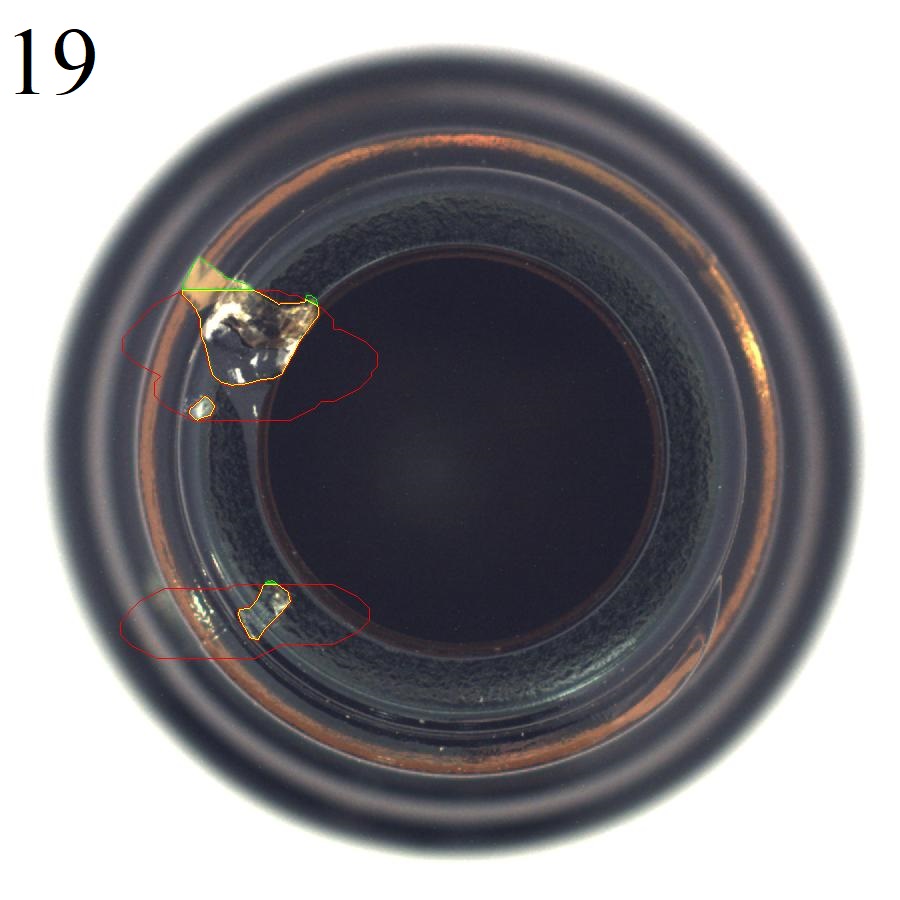}
		\end{subfigure}%
		\begin{subfigure}[h]{0.23\columnwidth}
			\includegraphics[width=\textwidth]{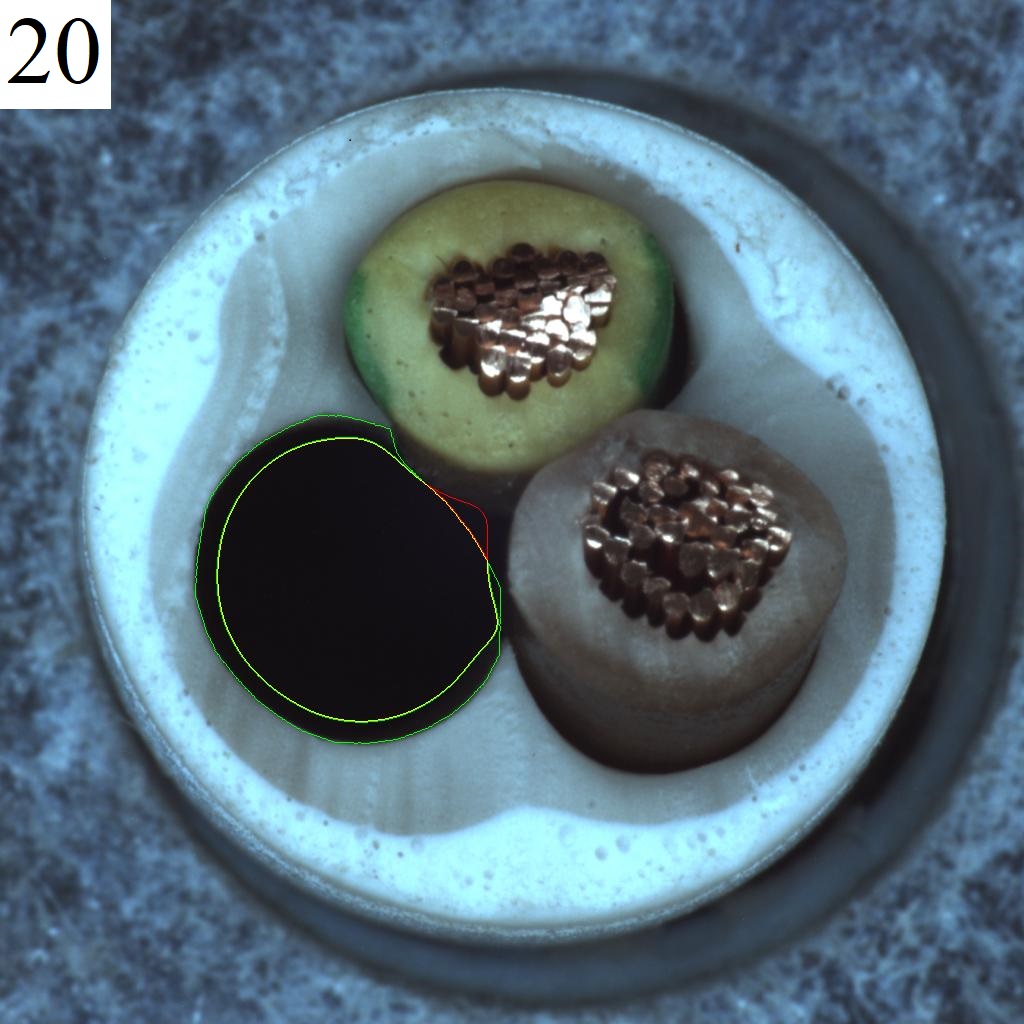}
		\end{subfigure}%
		\begin{subfigure}[h]{0.23\columnwidth}
			\includegraphics[width=\textwidth]{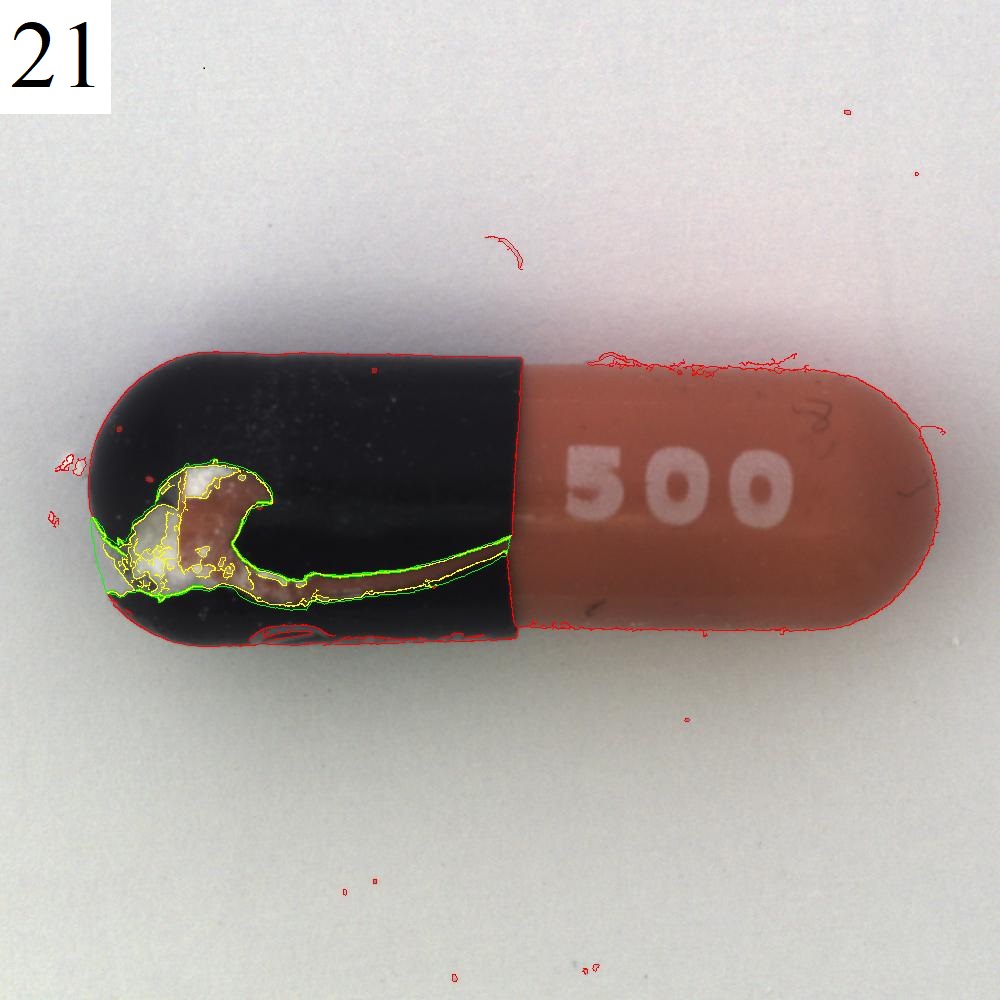}
		\end{subfigure}%
		\begin{subfigure}[h]{0.23\columnwidth}
			\includegraphics[width=1.0\textwidth]{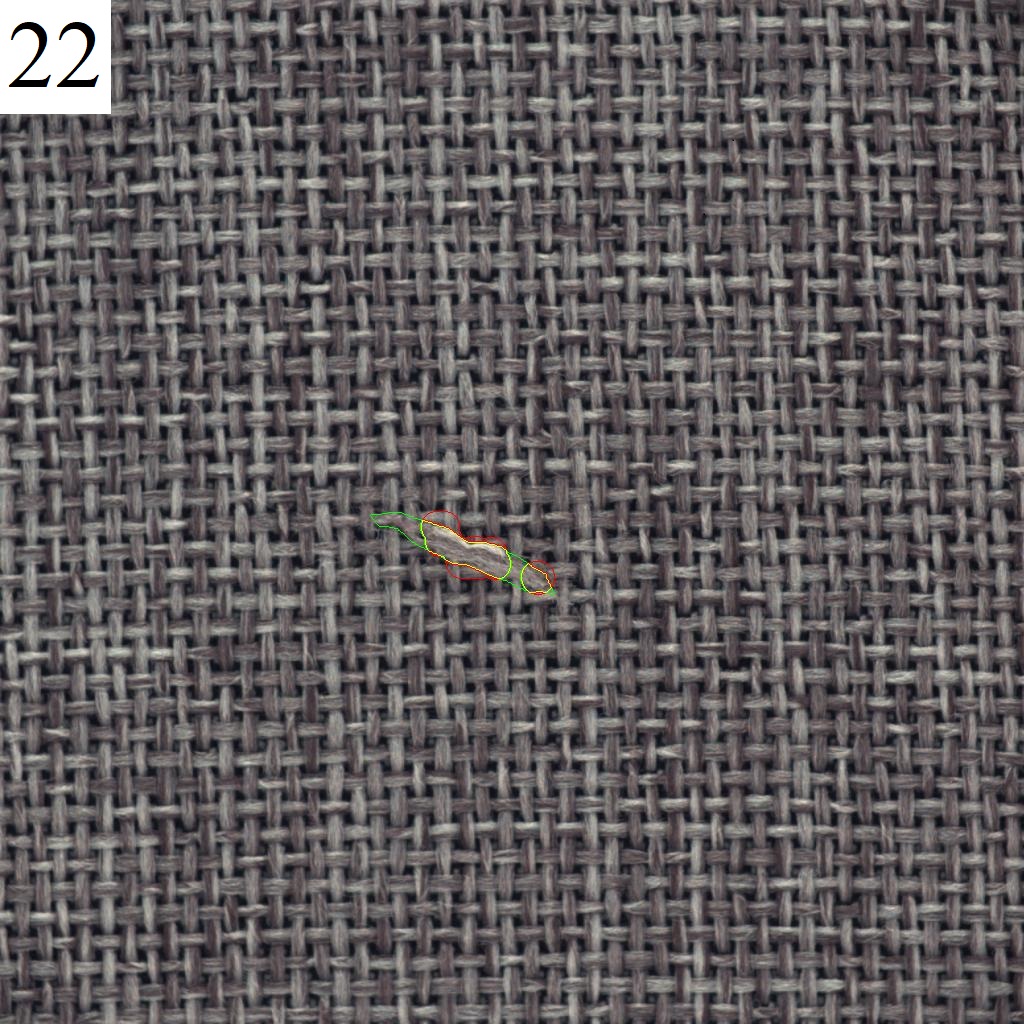}
		\end{subfigure}
		~
		\begin{subfigure}[h]{0.23\columnwidth}
			\includegraphics[width=\textwidth]{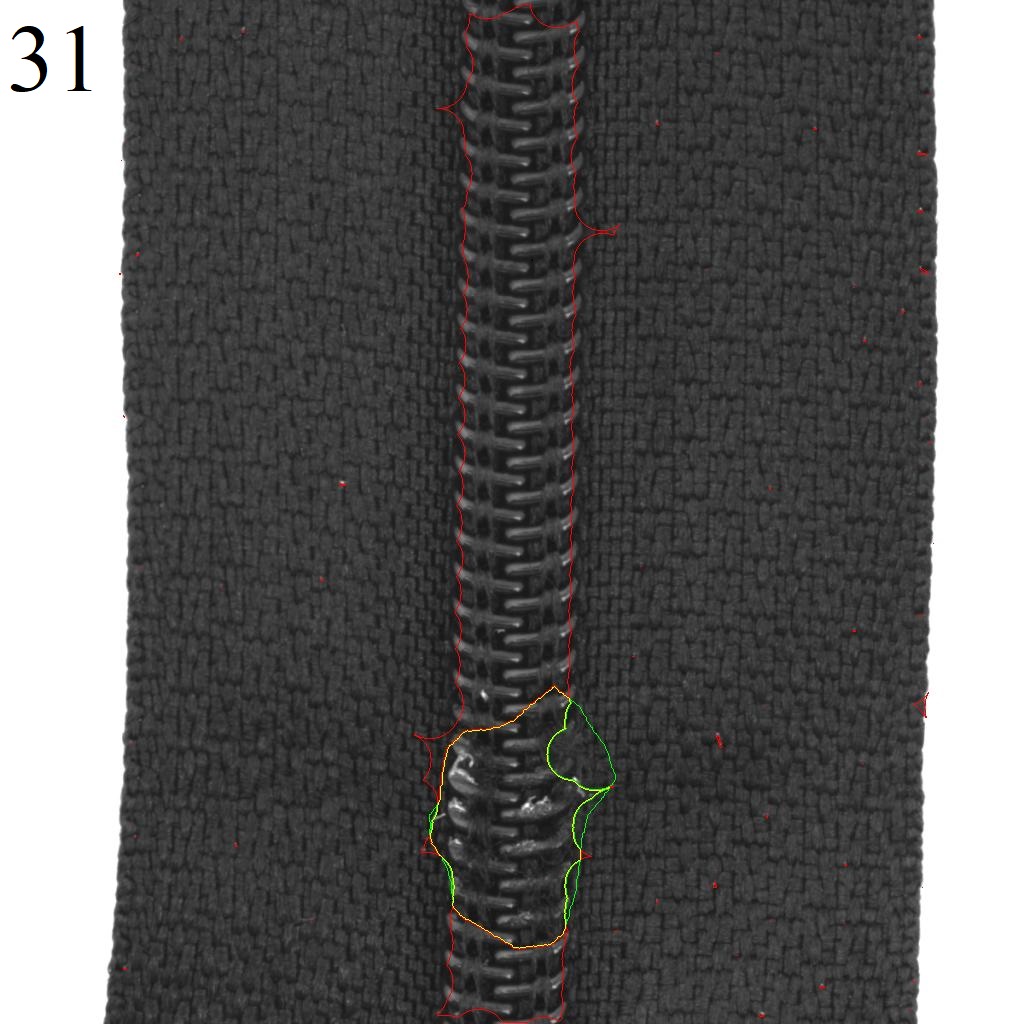}
		\end{subfigure}%
		\begin{subfigure}[h]{0.23\columnwidth}
			\includegraphics[width=\textwidth]{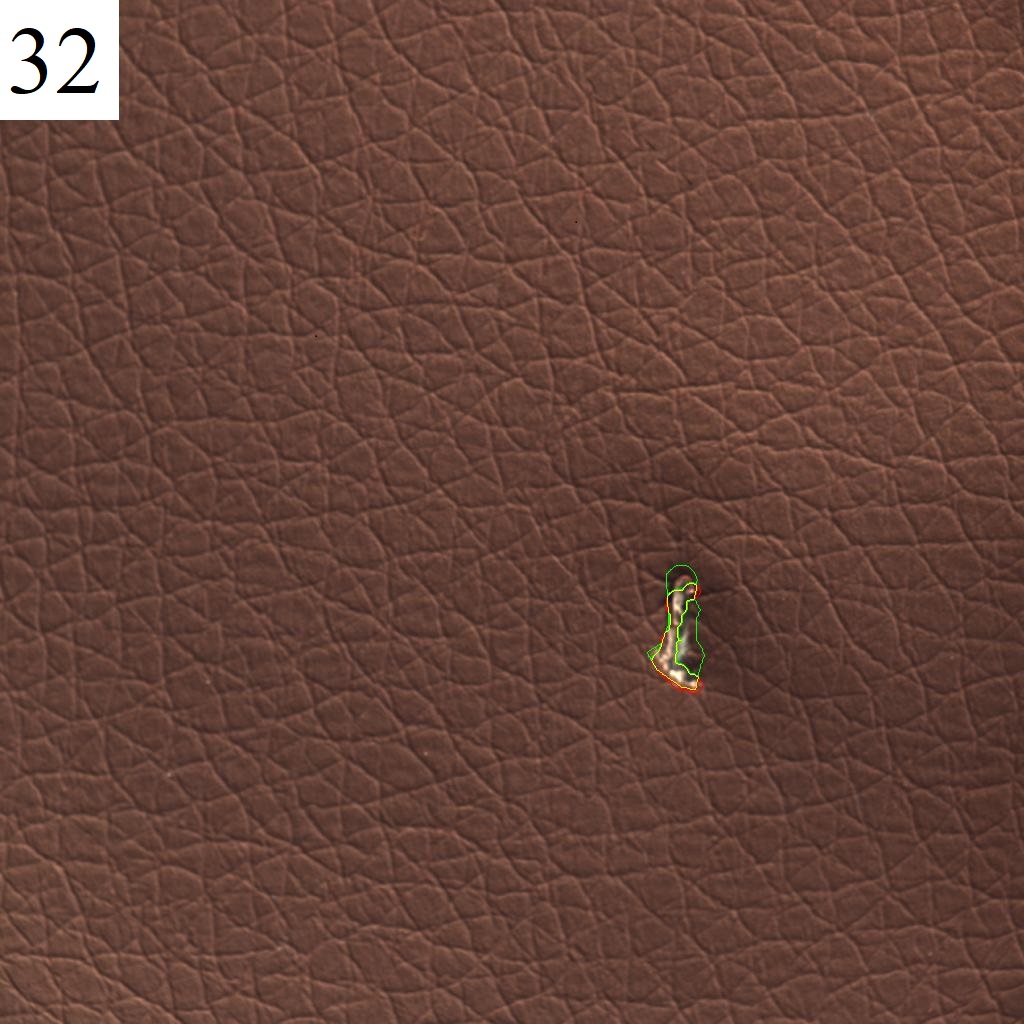}
		\end{subfigure}%
		\begin{subfigure}[h]{0.23\columnwidth}
			\includegraphics[width=1.0\textwidth]{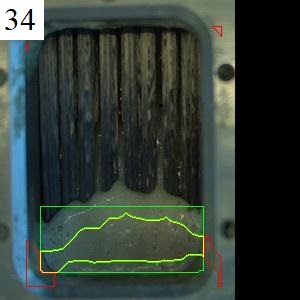}
		\end{subfigure}%
		\begin{subfigure}[h]{0.23\columnwidth}
			\includegraphics[width=\textwidth]{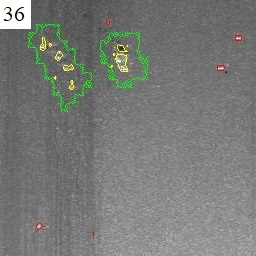}
		\end{subfigure}
		\caption{12 Samples out of 38 with labeled surface defects (green: ground truth, yellow: true positive, red: false positive) which give an insight into the data collection used for this study; based on previous work by Margraf et al. \cite{Margraf2023a,Margraf2025}}
		\label{fig:segmentation_datasets}
	\end{figure}
	
	\subsection{Similarity Distribution in Referenced Industrial Datasets}
	For the proof of concept, we selected images from industrial monitoring setups that were partly created in lab settings (labeled 01 - 10, 14 - 17 according to Tab. \ref{tab:datasets} on p. \pageref{tab:datasets}). Visualizations of some surface samples can be found in Fig. \ref{fig:segmentation_datasets} on p. \pageref{fig:segmentation_datasets}. The images show details from the following material:  
	Fuzzballs on nonwoven fabric (01-10), knitting anomalies on carbon fibers (14 -17), welding (36)\cite{Severstal2019} and metal defects (12,13,36) \cite{Huang2020}, MVTec Anomaly Detection Dataset (18-32,37)\cite{bergmann2019} and others (11,38).
	
	\begin{table}[h]
		\centering
		\begin{tabular}{ccccccc}
			\hline
			\textbf{Level} & \textbf{CNN} & \textbf{HE} & \textbf{TC} & \textbf{ED} & \textbf{SP} & \textbf{FF}\\
			\hline
			None & $61.01\%$ & $0.14\%$ & $0\%$ & $0.14\%$ & $0\%$ & $0\%$\\
			Low & $23.04\%$ & $0\%$ & $0\%$ & $0\%$ & $0\%$ & $0\%$\\
			Med & $8.53\%$ & $1.99\%$ & $3.41\%$ & $3.41\%$ & $12.52\%$ & $0.14\%$\\
			High & $5.41\%$ & $97.78\%$ & $96.59\%$ & $96.44\%$ & $87.48\%$ & $99.86\%$\\
			\hline
		\end{tabular}
		\caption{Value distribution by different similarity metrics for selected datasets, split into levels of similarity $s$: None ($s <= 0.3$), Low ($0.3 < s <= 0.5$), Medium ($0.5 < s <= 0.7$) and High ($0.7 < s <= 1.0$)}
		\label{tab:dataset_distribution}
	\end{table}
	
\begin{figure}[t!]
		\centering
		\includegraphics[width=1.0\columnwidth]{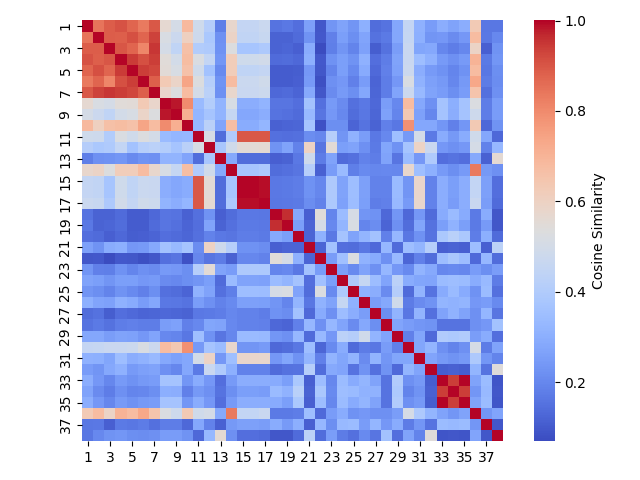}
		\caption{Cosine similarity computed by CNN embeddings between all pairs of datasets with dataset indices on both axis (cf. Appdx, Tab. \ref{tab:datasets}); the lowest similarities are reported in blue, highest in red for a value range of $[0.0,1.0]$}\label{fig:cosine_corr}
	\end{figure}

	Tab. \ref{tab:dataset_distribution} lists the distribution of similarities computed on 38 datasets. This distribution differs significantly among the 1,444 pairings, as can be seen in the heatmap for ResNet embeddings depicted in Fig. \ref{fig:cosine_corr}.
	The majority of CNN-based ResNet value pairs fall into the lower ranges, with 61.01\% classified as having very low or no similarity ($s \leq 0.3$) and an additional 23.04\% in the low similarity range. This indicates that most datasets are highly distinct in terms of their learned feature representations.
	In contrast, the remaining metrics (HE, TC, ED, SP, FF) show a different behavior, with the vast majority of dataset pairs concentrated in the high similarity range (between 87.48\% and 99.86\%). Only a tiny fraction of pairs fall into the none or low similarity categories for these metrics.
	
	Medium similarity levels occur only sparsely across all metrics, with the highest proportion observed for superpixel-based similarity (12.52\%), while for most other metrics this range remains below 5\%. Overall, the results suggest that CNN-based similarity provides a much stronger discriminative separation between datasets, whereas histogram, edge-based or statistical features tend to indicate uniformly high similarity across dataset pairs.
	
	However, the pairs of highly similar datasets are of particular interest. Due to their shared feature space, they represent the promising candidates for cross-application of evolved processing pipelines. We hypothesize that high-similarity pairs are more likely to allow for successful pipeline cross-application compared to low- or medium-similarity pairs.
	
	\subsection{Statistical Analysis of Cross-Application by Similarity}
	In the following sections, two sets of pipeline cross-application have been selected per dataset for closer examination:
	\begin{itemize}
		\item Pipelines closest to the \textit{Mean MCC} (MEAN)
		\item Pipelines offering the \textit{Best MCC} (BEST)
	\end{itemize}
	
\begin{figure*}[t!]
		\centering
		\begin{subfigure}{0.32\textwidth}
			\centering
			\includegraphics[width=\linewidth]{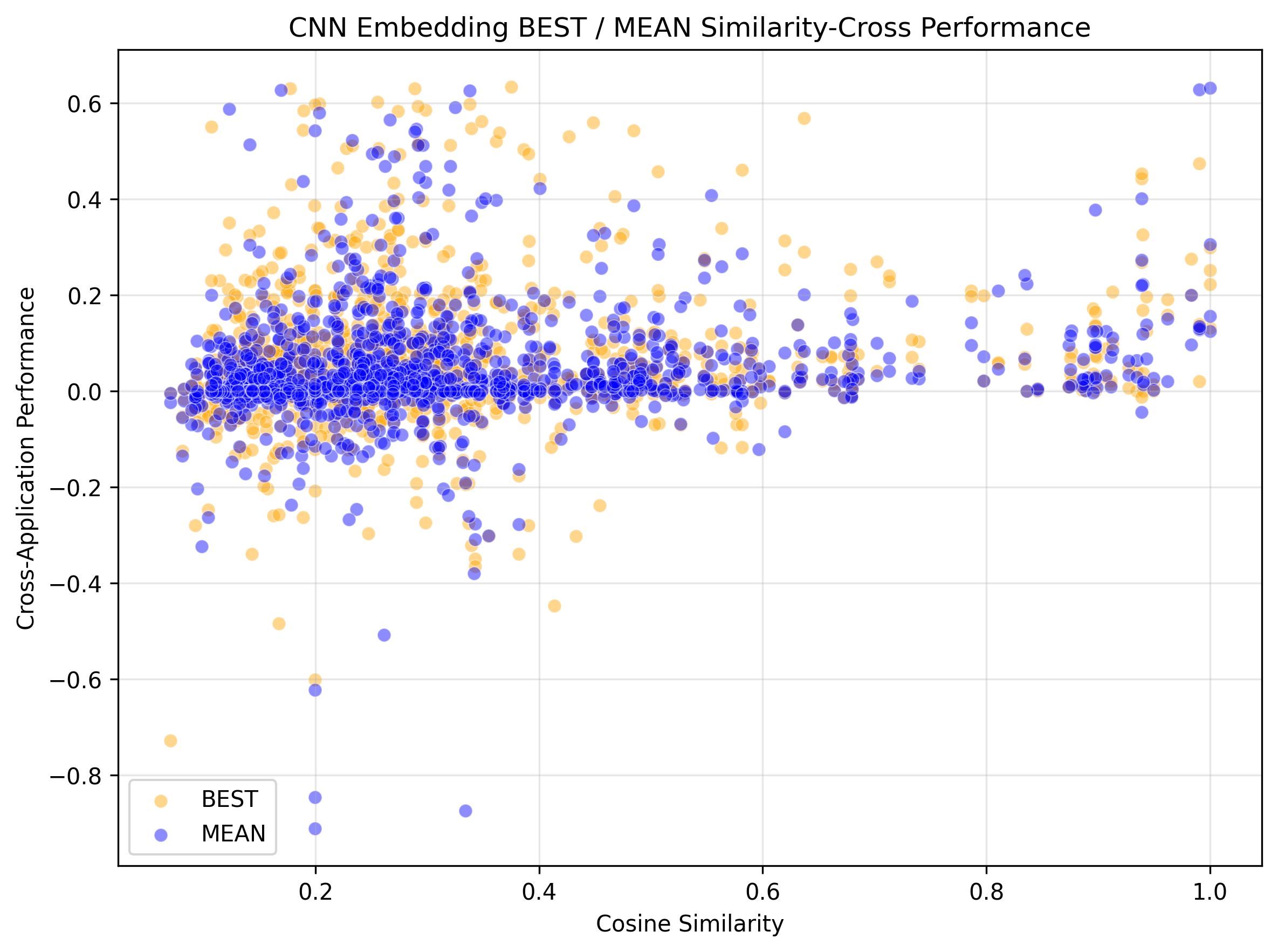}
		\end{subfigure}
		\hfill
		\begin{subfigure}{0.32\textwidth}
			\centering
			\includegraphics[width=\linewidth]{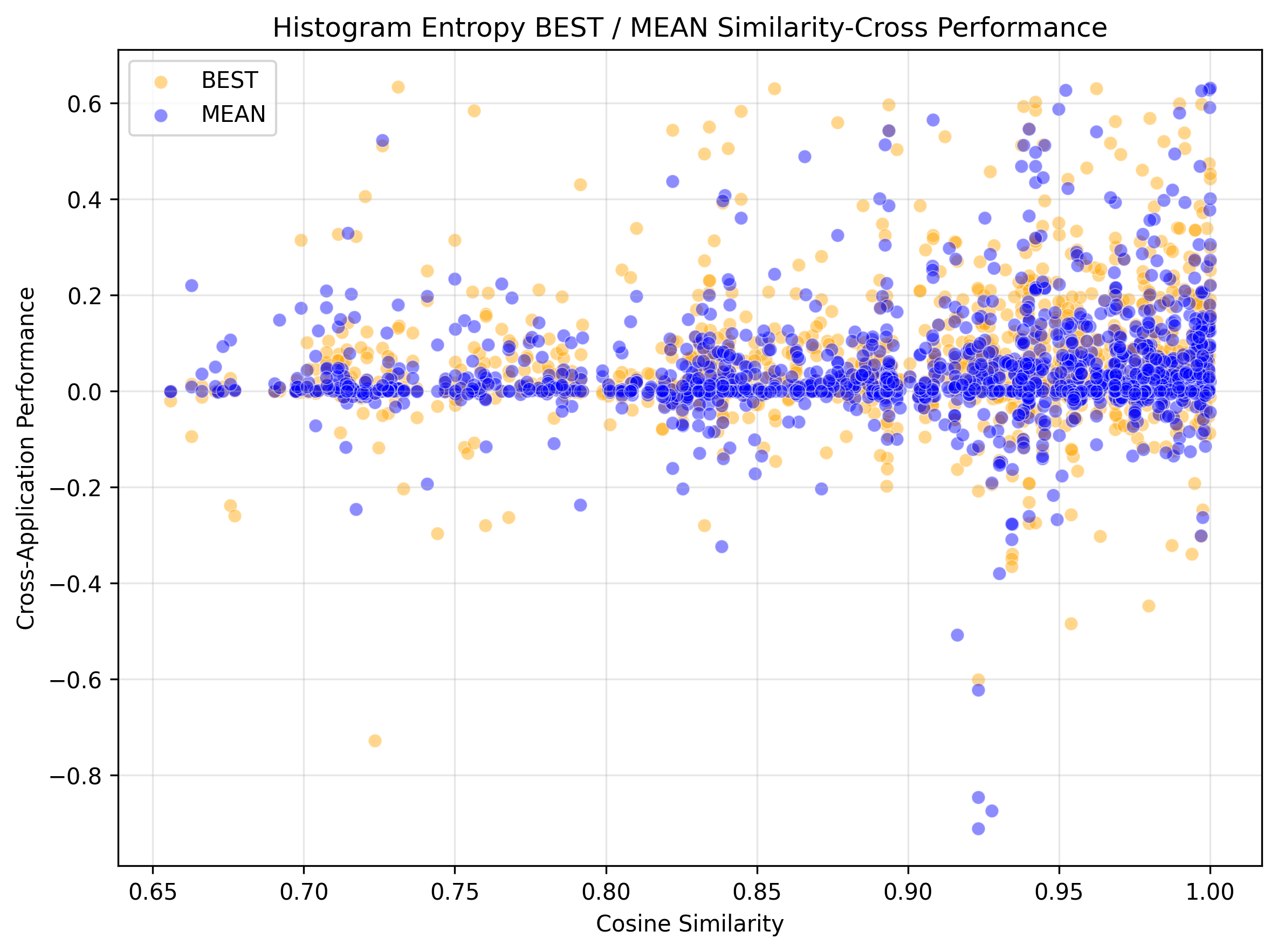}
		\end{subfigure}
		\hfill
		\begin{subfigure}{0.32\textwidth}
			\centering
			\includegraphics[width=\linewidth]{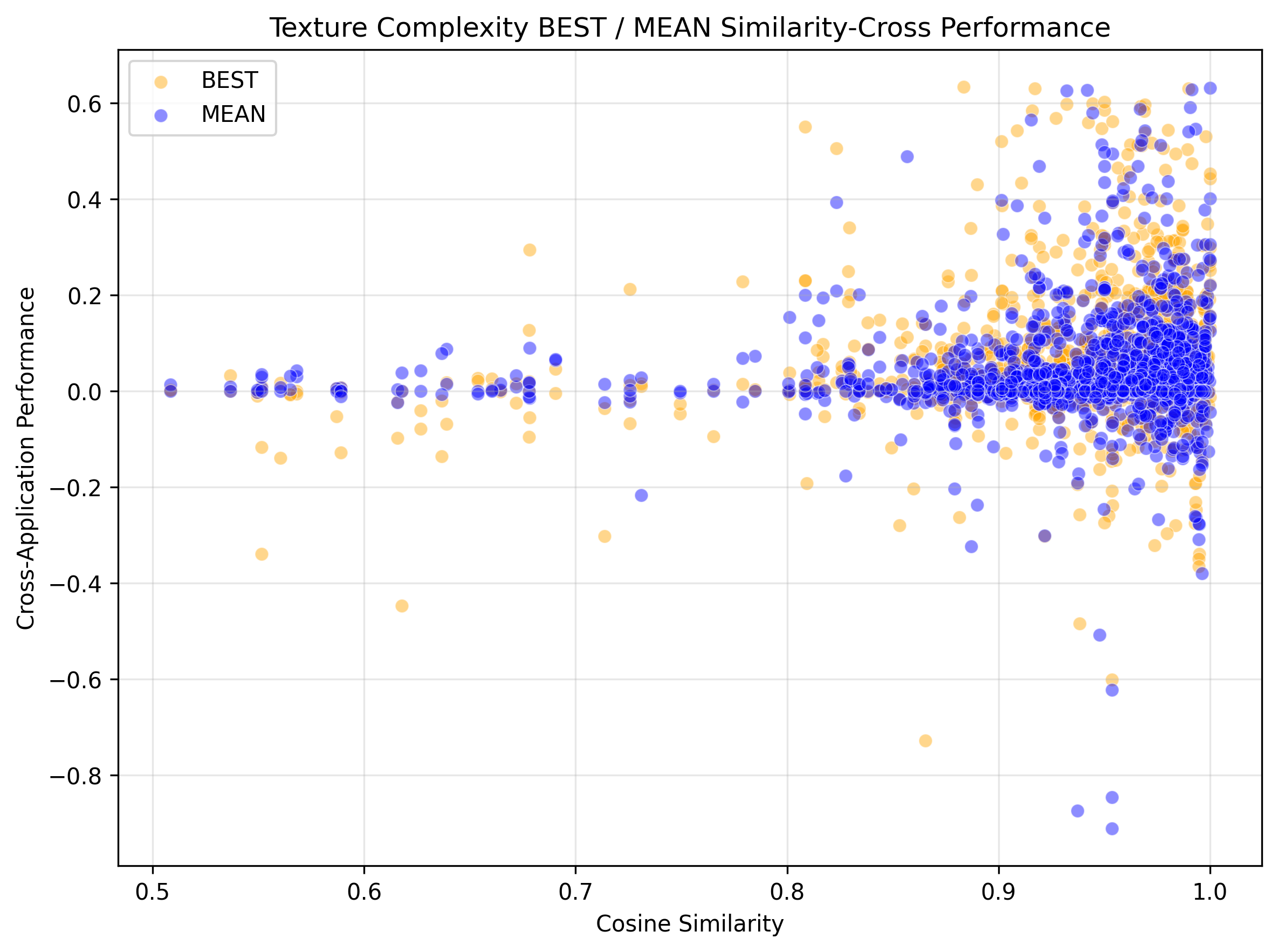}
		\end{subfigure}
		\caption{Regression scatterplots for dataset similarity metrics and cross-application performance of  MEAN/BEST; Left: CNN, Center: HC, Right: TC}
		\label{fig:similarity_cross_score}
	\end{figure*}

	\paragraph{Correlation Analysis}
	The correlation analysis between all 38 datasets yields a Pearson correlation of $r \in [0.077, 0.153]$ and a Spearman $\rho \in [0.048, 0.220]$ for \textit{MEAN}, as well as $r \in [0.102, 0.143 ]$ and $\rho \in [0.09, 0.167]$ for \textit{BEST} on all complexity metrics, cf. Tab. \ref{tab:single_metric_results}, p. \pageref{tab:single_metric_results}. Although these values indicate a poor overall correlation, the view slightly improves in the bin-wise analysis. The best values in Tab. \ref{tab:correlation_bins} on p. \pageref{tab:correlation_bins} are highlighted in bold letters. As can be seen, CNN embeddings reveal a higher correlation of $r = 0.292$ and $\rho = 0.315$ while TC even reports values close to $0.5$. It can be concluded, that depending on the complexity metric, there is a dependency between image complexity and pipeline performance. This is supported by the value distribution shown in Fig. \ref{fig:similarity_cross_score} on p. \pageref{fig:similarity_cross_score}.
	
	\begin{table}[h]
		\centering
		\begin{tabular}{llcccc}
			\toprule
			\textbf{MCC} & \textbf{Metric} & \textbf{Bin} & \textbf{$r$} & \textbf{$\rho$} & \textbf{$p$-value range} \\
			\midrule
			Mean & CNN        & $[0.8,1.0]$ & \textit{0.292} & \textit{0.315} & $<0.05$ \\
			Mean & HE    & $[0.8,1.0]$ & 0.117 & 0.146 & $<0.001$ \\
			Mean & FF  & $[0.8,1.0]$ & 0.030 & 0.043 & n.s. \\
			Best & SP & $[0.8,1.0]$ & 0.009 & -0.005 & n.s. \\
			Best & TC    & $[0.6,0.7]$ & \textbf{0.498} & \textbf{0.469} & $<0.05$ \\
			\bottomrule
		\end{tabular}
		\caption{Bin-wise correlation analysis for Pearson $r$ and Spearman $\rho$ between similarity and transfer performance. Only similarity bins (Bin) with sufficient observations are shown. Highest values are shown in bold, second-highest in italics.}\label{tab:correlation_bins}
	\end{table}
	
	The scatterplots in Fig. \ref{fig:similarity_cross_score} (p. \pageref{fig:similarity_cross_score}) show the dependency complexity versus cross-pipeline MCC performance for the most promising complexity metrics: CNN, HE, and TC.
	
	CNN indicates a concentration at moderate complexity values (0.0–0.4) and cross-pipeline performance (MCC) between $-0.4$ and $0.6$. HE exhibits a more uniform distribution with high complexity concentrations around $0.9 - 1.0$ but only weak dependency to cross-pipeline MCC. TC is mainly concentrated at high complexity values ($0.8 - 1.0$) with moderate MCC variation.
		
	\paragraph{Single and Multi-metric Logistic Regression Analysis}
	All models were fitted on $n=1406$ source--target transfer observations. 
	As the results listed in Tab. \ref{tab:single_metric_results} show, the strongest \textit{single} predictor for \emph{MEAN} transfer performance is CNN-based similarity ($R^2=0.023$), followed by HE ($R^2=0.013$). For \emph{BEST} transfer performance, SP and TC similarity yield the largest single-metric explanatory power ($R^2=0.020$ and $R^2=0.019$, respectively).	
	\begin{table}[h]
		\resizebox{\columnwidth}{!}{%
			\begin{tabular}{l c c c c c c}
				\toprule
				& \multicolumn{3}{c}{\textbf{Mean MCC}} & \multicolumn{3}{c}{\textbf{Best MCC}} \\
				\cmidrule(lr){2-4}\cmidrule(lr){5-7}
				\textbf{Sim} & {$r$} & {$\rho$} & {$R^2$} & {$r$} & {$\rho$} & {$R^2$} \\
				\midrule
				CNN         & \textbf{0.153} & \textbf{0.220} & \textbf{0.023} & 0.124 & 0.154 & 0.015 \\
				ED        & 0.068 & 0.105 & 0.005 & 0.116 & \textbf{0.167} & 0.013 \\
				TC     & 0.080 & 0.095 & 0.006 & 0.137 & 0.125 & \textbf{0.019} \\
				HE     & 0.112 & 0.151 & 0.013 & 0.107 & 0.131 & 0.011 \\
				FF   & 0.037 & 0.048 & 0.001 & 0.102 & 0.141 & 0.010 \\
				SP  & 0.077 & 0.079 & 0.006 & \textbf{0.143} & 0.090 & \textbf{0.020} \\
				\midrule
				\textbf{Best metric} & \multicolumn{3}{c}{CNN} & \multicolumn{3}{c}{Superpixel / Texture} \\
				\bottomrule
			\end{tabular}%
		}	
		\caption{Association between similarity metrics and cross-dataset transfer performance. Reported are Pearson correlation $r$, Spearman rank correlation $\rho$, and the coefficient of determination $R^2$ from separate single-metric OLS models. Highest values are shown in bold.}
		\label{tab:single_metric_results}
	\end{table}
	Most correlations appeared statistically significant; their effect sizes, however, remained comparably small. This finding indicates that image similarity allows to make assumptions on pipeline performance, but it may not suffice as an exclusive metric. The data suggest that although there is a significant dependency, it explains a limited fraction of transferability.
	
	The \textit{multi-metric} regression models shown in Tab.~\ref{tab:multimetric_models} quantify the influence of each similarity descriptor while controlling for the remaining predictors. The coefficients (Coef.) indicate the magnitude of the corresponding effect with p-values revealing their significance.		
	CNN and HE similarity appear as the most consistent predictors across all variants with positive coefficients and significant p-values for \emph{MEAN}, \emph{BEST}, and transferability. This suggests that visually similar datasets tend to yield better transfer performance. SP similarity stands out for \emph{BEST} and transferability which is a strong indicator that superpixel-dependent image structures hold information that allows to identify 'promising' pipelines.
		
	In contrast, ED, TC, and FF show limited explanatory power with coefficients remaining small and mostly insignificant.
	Interestingly, the coefficient of the original score is negative for all models. This suggests that strong performance on the source dataset alone does not necessarily translate into high cross-dataset transferability and even confirm the assumption that pipelines are highly optimized for their specific dataset in the first place.
	
	\begin{table}[h]
		\centering
		\resizebox{\columnwidth}{!}{%
			\begin{tabular}{l c c c c c c}
				\toprule
				& \multicolumn{2}{c}{\textbf{Mean MCC (OLS)}} 
				& \multicolumn{2}{c}{\textbf{Best MCC (OLS)}} 
				& \multicolumn{2}{c}{\textbf{Transferability (Logit)}} \\
				\cmidrule(lr){2-3}\cmidrule(lr){4-5}\cmidrule(lr){6-7}
				\textbf{Predictor} & {Coef.} & {$p$ range} & {Coef.} & {$p$ range} & {Coef.} & {$p$ range} \\
				\midrule
				CNN            &  0.015 & $<$0.001 &  0.013 & $<$0.001 &  0.178 & 0.003 \\
				ED           &  0.004 &  0.319 &  0.003 &  0.583 &  0.155 & 0.076 \\
				TC        &  0.002 &  0.608 &  0.004 &  0.408 &  0.169 & 0.059 \\
				HE        &  0.013 & $<$0.001 &  0.010 &  0.014 &  0.219 & 0.002 \\
				FF      & -0.006 &  0.158 &  0.004 &  0.361 &  0.020 & 0.823 \\
				SP     &  0.008 &  0.059 &  0.016 & $<$0.001 &  0.255 & 0.002 \\
				Original score & -0.005 &  0.147 & -0.009 &  0.017 & -0.120 & 0.049 \\
				\midrule
				Model fit       & \multicolumn{2}{c}{$R^2=0.040$} 
				& \multicolumn{2}{c}{$R^2=0.050$}
				& \multicolumn{2}{c}{Pseudo-$R^2=0.052$} \\
				Observations    & \multicolumn{2}{c}{1406}
				& \multicolumn{2}{c}{1406}
				& \multicolumn{2}{c}{1406} \\
				\bottomrule
			\end{tabular}%
		}
		\caption{Compact summary of multivariate regression models for transferability with estimated regression coefficient ($\beta_k$ / Coef.) and p-values.}
		\label{tab:multimetric_models}
	\end{table}
	It should be noted that the explanatory power of the models remains modest ($R^2=0.040$ for \emph{MEAN}, $R^2=0.050$ for \emph{BEST}, and pseudo-$R^2=0.052$ for the logistic model).	
	Hence, the proposed similarity descriptors explain only parts of the transferability between datasets. Cross-dataset CGP generalization likely depends not only on global image similarity characteristics, but also on additional structural, semantic, or optimization-related factors that are not explicitly covered in the present analysis.
	
	\subsection{Summary and Interpretation}
	
	This second analysis demonstrates a moderate improvement of \textit{BEST} over \textit{MEAN} pipelines, while highlighting fundamental differences between peak and average transfer performance. As expected, cross-application of pipelines tends to yield higher initial MCC values with increasing dataset similarity, although the overall effect remains limited.
	
	The results suggest that similarity can provide a weak but useful indicator for pipeline reuse and particularly for identifying promising starting points for evolutionary computation. This may help reduce unnecessary exploration and shorten early optimization phases. However, given the moderate overall explanatory power of the models, similarity alone is insufficient to reliably predict transfer performance, and additional dataset characteristics or meta-features are required to further improve efficiency. Future research is required to explore this direction.
	
	\subsection{Ressource-efficient Algorithm Scaling}
	The concept of retrieving data processing pipelines from a database by similarity - as proposed in this study - addresses industrial application scenarios on specific detection problems that cannot be solved by collecting arbitrary amounts of data. 
	For real-world scenarios that require a highly specialized solution based on data not available `in the wild', an approach is required to build solutions even when data sources are limited.
	Furthermore, sensitive applications with large data streams become inefficient if high resolution signal or image data has to be processed within narrow time windows.
	
\begin{figure}[h]
		\centering
		\includegraphics[width=1.0\columnwidth]{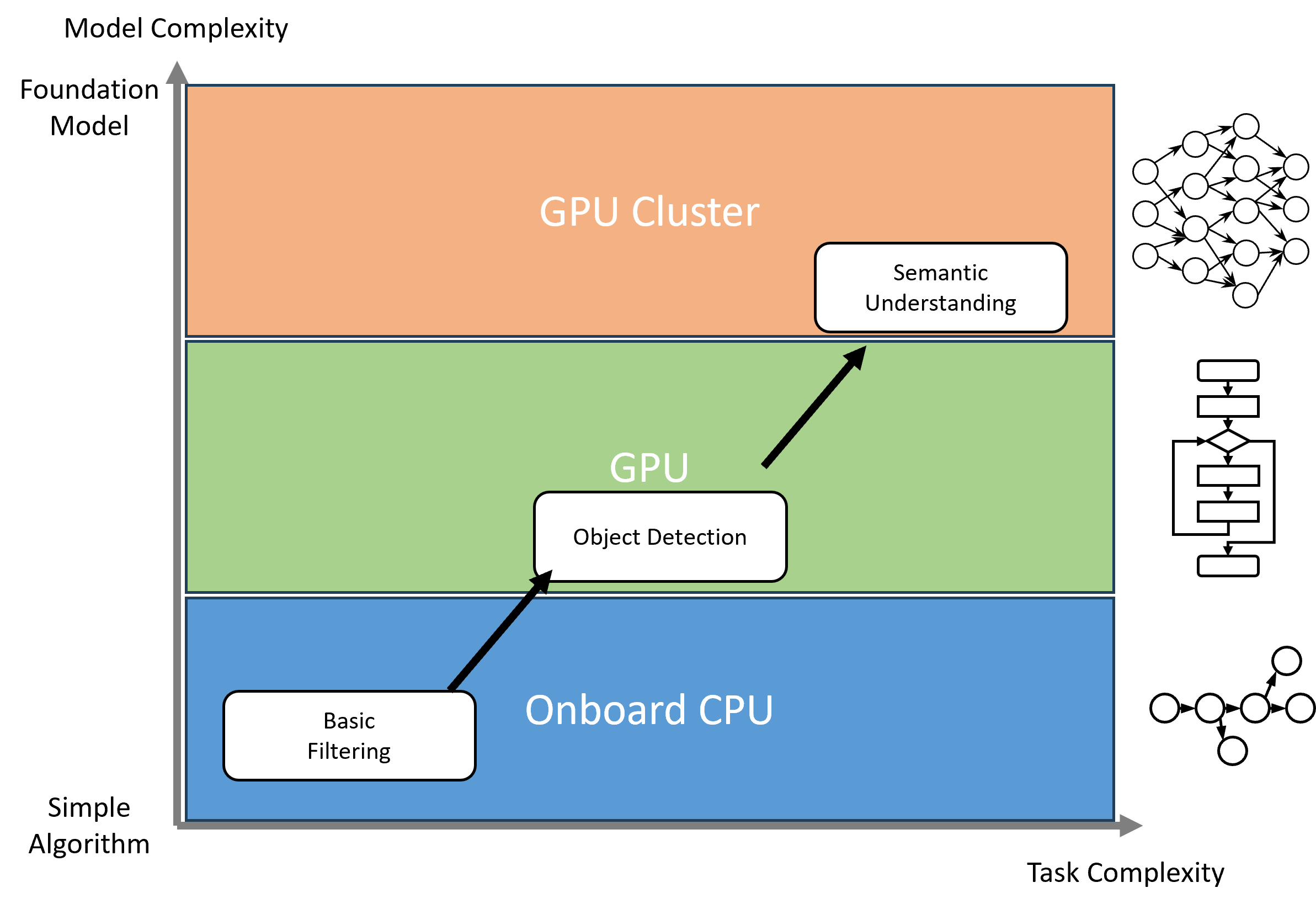}
		\caption{Architecture scaling according to algorithms complexity and computational needs: We distinguish between 3 major types of processor units - \textit{FPGA}, \textit{GPU} and \textit{CPU}.}\label{fig:arch_selection}
	\end{figure}

	As illustrated in Fig. \ref{fig:arch_selection}, the task complexity in combination with annotation features and high-level, abstract image or signal properties can be utilized to determine the best suited computing architecture. On the basis of this relation, a classifier model can be designed and trained to specifically match the most efficient architecture. These classifiers consist of, but are not limited to, the following choices: 
	\textit{learning-classifier systems}, \textit{random forests}, \textit{support vector classifier}, \textit{CNN}, \textit{RNN}, \textit{Transformers}.
	The output should then return suggestions for the following design parameters: the computing system (pc, edge device), the processing unit (CPU, GPU, FPGA) and the context-related classification problem (object detection, image classification, semantic/panoptic segmentation).
	
	We want to point out, that specialized monitoring systems depend more on stability and reproducibility than approximate and variable environment detection abilities. Also, most applications exhibit different levels of difficulty and complexity (cf. ~\cite{Geinitz2016,Margraf2017a,Mertes2022}). Learning to detect a complicated, extremely variable defect (e.g. scratches, object dislocations, foreign material or dirt) represents a far more difficult task to accomplish. Fig. \ref{fig:arch_selection} shows how the architecture design parameters must be defined according to the complexity of the task. In addition, the quality of a model output depends of dataset size and quality: However, collecting sufficient amounts of labels and of sufficient quality imposes a time-consuming task. With a decreasing amount of annotated data, but increasing limitation and definition of the detection scenario, filter algorithms tend to outperform competing deep learning models. The latter is a very likely scenario in many industrial use cases. This insight can be broken down to: \textit{Simple problems call for simple solutions} and \textit{the right tool should be applied to the right job}.
	
	In this study, we examined how small, efficient programs allow for a better performance on low complex tasks in highly specialized environments. We still recommend complex model architectures for tasks of higher complexity but also emphasize their drawbacks: Large foundation models only use their potential if one model is required to meet a wide range of applications with a high level of transferability and high abstraction, e.g. code generation or one-shot image classification. However, this comes with a higher risk of inaccuracies that are difficult to evaluate, an effect known as `hallucination'. This risk can be mitigated more effectively by our approach which relies on small, but highly specialized models using methodologies of EC. 	
	In general, our approach is intended to reduce design time and speed up development of monitoring systems in industrial applications. Ultimately, the concept presented in this study will allow for more efficient engineering solutions and reduce cost and time effort. 
	
	\section{Discussion}\label{sec:discussion}
	The similarity scores that were computed for selected image data from real-world environments (cf. Fig. \ref{fig:segmentation_datasets}, Tab. \ref{tab:datasets}) show similarities that, for one part, appear to confirm in parts the initial assumptions: Similar structures yield higher similarity for e.g. knitted fibers or nonwoven carbon fiber (CF, MP2.0) which appear to emphasize common characteristics between the sample and the reference images.
	
	The pipelines that were previously evolved using CGP for earlier studies were cross-applied between all datasets. The following performance comparison shows moderate improvement in best MCC over pipelines with a value close to the mean.
	However, the promising experimental runs call for further examination in the field. The extraction of additional image features will likely allow for a more specific search on similar inspection tasks.
	
	Although the approach appears to partly support the hypothesis that the retrieval of pipelines harnesses CGP-based solutions, we propose to a) extend the metrics used to estimate the fit of the CGP algorithm on image segmentation tasks and b) apply additional features beyond image similarity to predict pipeline performance. Dividing the problems into smaller sets of segmentation subtasks will likely increase the efficiency and performance of a pipeline. For industrial textiles, this would equal a preprocessing step that extracts the fiber within the visible area before the search for more complex anomalies such as gaps, holes or misaligned fibers begins. With 1,406 unique evaluations collected over two rounds for mean and best MCC values, this is the largest study in this industry-oriented field so far.
	
	Furthermore, the authors argue that a well-selected set of parameters allows for accurate predictions on system design. If we consider image size, the number of color channels, and histogram entropy, we should be able to identify a good fit for the overall system architecture that contains: Processing and filter pipeline, classifier model, and processor architecture. By closely analyzing the input data and the underlying task, system engineers and machine learning practitioners will be able to select filter algorithms and classification models with respect to the application domain. The authors of this paper would like to encourage AI engineers to critically reflect their expectations towards a model's classification performance and generalizability. Keeping in mind the computational resources and dataset size, the models should be as complex as necessary, but simple if applicable. Of course, this is a fine line, and a change of external requirements bears the risk of extensive reconfiguration. However, the methodology presented in this paper explicitly allows for adaptations of system architecture especially in low-power, low-latency, and resource-efficient environments.
	
	\section{Concluding Remarks}\label{sec:conclusion}
	This paper continues the complexity analysis on image datasets by Margraf et al. \cite{Margraf2017a,Margraf2023a,Margraf2025} and extends it by investigating the reuse of filter pipelines in industrial image processing. We examined whether pipelines, once generated and optimized for a particular dataset, can be efficiently retrieved from a database and adapted for use on similar datasets with minimal modification. This particular approach has been inspired by the fact that training time and effort can be reduced by reusing and fine-tuning existing pipelines, rather than evolving new pipelines without any prior knowledge. A similar concept has been used for CNN model training before where it is denoted transfer learning and has ever since become a standard methodology.
	
	In summary, we proposed a three-tier methodology that uses similarity scores from data sets to retrieve and cross-apply on pipelines. This is then followed by a statistical analysis on performance. The results presented in this study indicate that while full retraining provides optimal performance, reusing pipelines with parameter optimization alone achieves a slight but statistically significant improvement over naive reuse. These findings provide evidence for the potential efficiency gains of pipeline reuse and supports the idea to build a dedicated collection of pipeline-dataset tuples for which they were originally developed.
	
	The study highlights the importance of understanding dataset characteristics and task complexity when designing quality assurance systems. As demonstrated in prior work, early design decisions can severely limit system flexibility and performance. Our results suggest that intelligent retrieval and reuse strategies can help balance flexibility and reliability, reduce manual effort, and lower the risk of costly design misconceptions. We recommend integrating such mechanisms into architecture selection workflows and regularly reviewing system configurations to maintain performance as operational conditions change over time.
	
	Although this work only considers a part of the possible combinations of datasets, tasks, and pipelines, it offers a promising foundation for simplifying the design of industrial monitoring systems. Future work should focus on refining similarity metrics, improving adaptive parameter tuning methods, and conducting evaluations on larger-scale datasets to improve the concept of pipeline retrieval and reuse.
	
	\footnotesize
	\bibliographystyle{IEEEtran}
	\bibliography{ref}

@InProceedings{Geinitz2016,
	author    = {Geinitz, Steffen and Margraf, Andreas and Wedel, Andr{\'e} and Witthus, Sebastian and Drechsler, Klaus},
	booktitle = {Proc. of 19th World Conference on Non-Destructive Testing},
	title     = {Detection of filament misalignment in carbon fiber production using a stereovision line scan camera system},
	year      = {2016},
	url       = {http://ndt.net/?id=19575},
}

@Article{Margraf2017a,
	author   = {Andreas Margraf and Steffen Geinitz and André Wedel and Leonhard Engstler},
	journal  = {SAMPE 2017},
	title    = {Detection of surface defects on carbon fiber rovings using line sensors and image processing algorithms},
	year     = {2017},
}

@inproceedings{Stein2018,
  title={Toward an organic computing approach to automated design of processing pipelines},
  author={Stein, Anthony and Margraf, Andreas and Moroskow, Juergen and Geinitz, Steffen and Haehner, Joerg},
  booktitle={ARCS Workshop 2018; 31th International Conference on Architecture of Computing Systems},
  pages={1--8},
  year={2018},
  organization={VDE}
}

@Article{Mertes2022,
  author      = {Silvan Mertes and Andreas Margraf and Steffen Geinitz and Elisabeth André},
  title       = {Alternative Data Augmentation for Industrial Monitoring using Adversarial Learning},
  year		  = {2022},
  eprint      = {2205.04222},
  eprintclass = {cs.CV},
  eprinttype  = {arXiv}
}

@Article{Halevy2009,
  author  = {Alon Halevy and Peter Norvig and Fernando Pereira},
  journal = {IEEE Intelligent Systems},
  title   = {The Unreasonable Effectiveness of Data},
  year    = {2009},
  pages   = {8-12},
  volume  = {24},
  url     = {http://www.computer.org/portal/cms_docs_intelligent/intelligent/homepage/2009/x2exp.pdf},
}

@inproceedings{Cui2023,
  title={Equidistant reorder operator for cartesian genetic programming},
  author={Cui, Henning and Margraf, Andreas and H{\"a}hner, J{\"o}rg},
  booktitle={IJCCI},
  pages={64--74},
  year={2023}
}

@inproceedings{Cui2024a,
  title={Positional bias does not influence cartesian genetic programming with crossover},
  author={Cui, Henning and Heider, Michael and H{\"a}hner, J{\"o}rg},
  booktitle={International Conference on Parallel Problem Solving from Nature},
  pages={151--167},
  year={2024},
  organization={Springer}
}

@inproceedings{Cui2024b,
  author    = {Henning Cui and J{\"o}rg H{\"a}hner},
  title     = {Cartesian Genetic Programming is robust against redundant attributes in datasets},
  booktitle = {Proceedings of the 16th International Joint Conference on Computational Intelligence - ECTA, November 20-22, 2024, in Porto, Portugal},
  editor    = {Francesco Marcelloni and Kurosh Madani and Niki van Stein and Joaquim Filipe},
  number    = {Volume 1},
  pages     = {108 -- 119},
  doi       = {10.5220/0012974600003837},
  year      = {2024},
}

@phdthesis{Margraf2025,
	author = {Margraf, Andreas},
	publisher = {University of Augsburg},
	title = {{Evolutionary Learining for Data Processing Pipelines in Industrial Monitoring}},
    year = {2026},
    pages = {177},  
    school = {University of Augsburg},
    type = {Doctoral Dissertation},
    url = {https://opus.bibliothek.uni-augsburg.de/opus4/130569}
}

@article{Hestness2017,
  title={Deep learning scaling is predictable, empirically},
  author={Hestness, Joel and Narang, Sharan and Ardalani, Newsha and Diamos, Gregory and Jun, Heewoo and Kianinejad, Hassan and Patwary, Md and Ali, Mostofa and Yang, Yang and Zhou, Yanqi},
  journal={arXiv preprint arXiv:1712.00409} ,
  year={2017}
}

@InBook{Harding2013,
	author    = {Simon Harding and Jürgen Leitner and Jürgen Schmidhuber},
	chapter   = {Cartesian Genetic Programming for Image Processing},
	editor    = {Riolo, Rick and Vladislaleva, Ekaterina and Moore, Jason H. and Ritchie, Marylyn D.},
	pages     = {1--17},
	publisher = {Kluwer Academic Publishers},
	title     = {Genetic programming theory and practice x},
	year      = {2013}
}

@Article{Pfisterer2018,
	author   = {Pfisterer, Florian and van Rijny, Jan N. and Probst, Philipp and Müller, Andreas and Bischl, Bernd},
	title    = {Learning Multiple Defaults for Machine Learning Algorithms},
	year     = {2018},
	doi      = {arXiv:1811.09409v1}
}

@Article{Kaufmann2013,
	author    = {Paul Kaufmann and Kyrre Glette and Thiemo Gruber and Marco Platzner and Jim Torresen and Bernhard Sick},
	journal   = {{IEEE} Transactions on Evolutionary Computation},
	title     = {Classification of Electromyographic Signals: Comparing Evolvable Hardware to Conventional Classifiers},
	year      = {2013},
	number    = {1},
	pages     = {46--63},
	volume    = {17},
	doi       = {10.1109/tevc.2012.2185845},
	publisher = {Institute of Electrical and Electronics Engineers ({IEEE})}
}

@Article{Kephart2003,
	author={Kephart, J.O. and Chess, D.M.},
	journal={Computer}, 
	title={The vision of autonomic computing}, 
	year={2003},
	volume={36},
	number={1},
	pages={41-50},
	doi={10.1109/MC.2003.1160055}
}

@Article{Hernandez2020,
	author      = {Danny Hernandez and Tom B. Brown},
	title       = {Measuring the Algorithmic Efficiency of Neural Networks},
	year		= {2020},
	date        = {2020-05-08},
	eprint      = {2005.04305},
	eprintclass = {cs.LG},
	eprinttype  = {arXiv}
}

@InProceedings{Yu2013,
	author    = {Honghai Yu and Stefan Winkler},
	booktitle = {2013 Fifth International Workshop on Quality of Multimedia Experience (QoMEX)},
	title     = {Image complexity and spatial information},
	year      = {2013},
	month     = {jul},
	publisher = {IEEE},
	doi       = {10.1109/qomex.2013.6603194}
}

@InCollection{Perkioe2009,
	author    = {Jukka Perkiö and Aapo Hyvärinen},
	booktitle = {Artificial Neural Networks {\textendash} {ICANN} 2009},
	publisher = {Springer Berlin Heidelberg},
	title     = {Modelling Image Complexity by Independent Component Analysis, with Application to Content-Based Image Retrieval},
	year      = {2009},
	pages     = {704--714},
	doi       = {10.1007/978-3-642-04277-5_71}
}

@Article{Ke2021,
	author    = {Rihuan Ke and Aurelie Bugeau and Nicolas Papadakis and Mark Kirkland and Peter Schuetz and Carola-Bibiane Schonlieb},
	journal   = {{IEEE} Transactions on Image Processing},
	title     = {Multi-Task Deep Learning for Image Segmentation Using Recursive Approximation Tasks},
	year      = {2021},
	pages     = {3555--3567},
	volume    = {30},
	doi       = {10.1109/tip.2021.3062726},
	publisher = {Institute of Electrical and Electronics Engineers ({IEEE})}
}

@inproceedings{Ionescu2016,
  title={How hard can it be? Estimating the difficulty of visual search in an image},
  author={Tudor Ionescu, Radu and Alexe, Bogdan and Leordeanu, Marius and Popescu, Marius and Papadopoulos, Dim P and Ferrari, Vittorio},
  booktitle={Proceedings of the IEEE Conference on Computer Vision and Pattern Recognition},
  pages={2157--2166},
  year={2016}
}

@Article{Ivanovici2020,
	author    = {Mihai Ivanovici and Radu-Mihai Coliban and Cosmin Hatfaludi and Irina Emilia Nicolae},
	journal   = {Journal of Imaging},
	title     = {Color Image Complexity versus Over-Segmentation: A Preliminary Study on the Correlation between Complexity Measures and Number of Segments},
	year      = {2020},
	month     = {mar},
	number    = {4},
	pages     = {16},
	volume    = {6},
	doi       = {10.3390/jimaging6040016},
	publisher = {{MDPI} {AG}}
}

@Article{Eggensperger2017,
	author      = {Katharina Eggensperger and Marius Lindauer and Frank Hutter},
	title       = {Pitfalls and Best Practices in Algorithm Configuration},
	year        = {2017},
	date        = {2017-05-17},
	eprint      = {1705.06058},
	eprintclass = {cs.AI},
	eprinttype  = {arXiv}
}

@Article{Domingos2012,
	author    = {Pedro Domingos},
	journal   = {Communications of the {ACM}},
	title     = {A few useful things to know about machine learning},
	year      = {2012},
	number    = {10},
	pages     = {78--87},
	volume    = {55},
	doi       = {10.1145/2347736.2347755},
	publisher = {Association for Computing Machinery ({ACM})}
}

@Article{Kerschke2018,
	author      = {Pascal Kerschke and Holger H. Hoos and Frank Neumann and Heike Trautmann},
	title       = {Automated Algorithm Selection: Survey and Perspectives},
	year        = {2018},
	date        = {2018-11-28},
	eprint      = {1811.11597},
	eprintclass = {cs.LG},
	eprinttype  = {arXiv},
}

@Book{muller2011organic,
	author    = {M{\"u}ller-Schloer, Christian and Schmeck, Hartmut and Ungerer, Theo},
	publisher = {Springer Science \& Business Media},
	title     = {Organic computing—a paradigm shift for complex systems},
	year      = {2011},
	isbn      = {9783034801300},
	series    = {Autonomic Systems}
}

@InProceedings{bergmann2019,
  author    = {Bergmann, Paul and Fauser, Michael and Sattlegger, David and Steger, Carsten},
  booktitle = {Proceedings of the IEEE/CVF conference on computer vision and pattern recognition},
  title     = {MVTec AD--A comprehensive real-world dataset for unsupervised anomaly detection},
  year      = {2019},
  pages     = {9592--9600},
}

@Misc{Severstal2019,
  title        = {Severstal Steel Dataset},
  author       = {Severstal},
  year         = {2019}, 
  howpublished = {\url{https://www.kaggle.com/c/severstal-steel-defect-detection}},
  note = {Accessed: 2022-05-03},
  url          = {https://www.kaggle.com/c/severstal-steel-defect-detection},
  urldate      = {2022-05-03},
}

@article{Huang2020,
  title={Surface defect saliency of magnetic tile},
  author={Huang, Yibin and Qiu, Congying and Yuan, Kui},
  journal={The Visual Computer},
  volume={36},
  number={1},
  pages={85--96},
  year={2020},
  doi={10.1109/coase.2018.8560423},
  publisher={Springer}
}

@inproceedings{Russakoff2004,
	title={Image similarity using mutual information of regions},
	author={Russakoff, Daniel B and Tomasi, Carlo and Rohlfing, Torsten and Maurer, Calvin R},
	booktitle={European Conference on Computer Vision},
	pages={596--607},
	year={2004},
	organization={Springer}
}

@inproceedings{Dalal2005,
  author={Dalal, N. and Triggs, B.},
  booktitle={2005 IEEE Computer Society Conference on Computer Vision and Pattern Recognition (CVPR'05)}, 
  title={Histograms of oriented gradients for human detection}, 
  year={2005},
  volume={1},
  number={},
  pages={886-893 vol. 1},
  doi={10.1109/CVPR.2005.177}
}

@InCollection{Redies2012,
  author    = {Christoph Redies and Seyed Ali Amirshahi and Michael Koch and Joachim Denzler},
  booktitle = {Computer Vision {\textendash} {ECCV} 2012. Workshops and Demonstrations},
  publisher = {Springer Berlin Heidelberg},
  title     = {{PHOG}-Derived Aesthetic Measures Applied to Color Photographs of Artworks, Natural Scenes and Objects},
  year      = {2012},
  pages     = {522--531},
  doi       = {10.1007/978-3-642-33863-2_54}
}

@article{Jain2020,
  author       = {Paras Jain and
                  Ajay Jain and
                  Aniruddha Nrusimha and
                  Amir Gholami and
                  Pieter Abbeel and
                  Kurt Keutzer and
                  Ion Stoica and
                  Joseph E. Gonzalez},
  title        = {Checkmate: Breaking the Memory Wall with Optimal Tensor Rematerialization},
  journal      = {CoRR},
  volume       = {abs/1910.02653},
  year         = {2019},
  url          = {http://arxiv.org/abs/1910.02653},
  eprinttype    = {arXiv},
  eprint       = {1910.02653},
  bibsource    = {dblp computer science bibliography, https://dblp.org}
}

@inproceedings{Sevilla2022,
   title={Compute Trends Across Three Eras of Machine Learning},
   url={http://dx.doi.org/10.1109/IJCNN55064.2022.9891914},
   DOI={10.1109/ijcnn55064.2022.9891914},
   booktitle={2022 International Joint Conference on Neural Networks (IJCNN)},
   publisher={IEEE},
   author={Sevilla, Jaime and Heim, Lennart and Ho, Anson and Besiroglu, Tamay and Hobbhahn, Marius and Villalobos, Pablo},
   year={2022},
   month=jul, pages={1–8} }

@misc{Goldblum2023,
      title={Battle of the Backbones: A Large-Scale Comparison of Pretrained Models across Computer Vision Tasks}, 
      author={Micah Goldblum and Hossein Souri and Renkun Ni and Manli Shu and Viraj Prabhu and Gowthami Somepalli and Prithvijit Chattopadhyay and Mark Ibrahim and Adrien Bardes and Judy Hoffman and Rama Chellappa and Andrew Gordon Wilson and Tom Goldstein},
      year={2023},
      eprint={2310.19909},
      archivePrefix={arXiv},
      primaryClass={cs.CV},
      url={https://arxiv.org/abs/2310.19909}, 
}

@InProceedings{Margraf2023a,
  author    = {Margraf, Andreas and Cui, Henning and Stein, Anthony and Hähner, Jörg},
  booktitle = {2023 IEEE International Conference on Autonomic Computing and Self-Organizing Systems (ACSOS)},
  title     = {Evolving Processing Pipelines for Industrial Imaging with Cartesian Genetic Programming},
  year      = {2023},
  month     = sep,
  publisher = {IEEE},
  doi       = {10.1109/acsos58161.2023.00031},
}
	
	\newpage

\normalsize

\appendices

\section{Metric Definitions and Dataset Labelling}
The similarity between datasets are computed as the cosine similarity of \textit{Histogram of Gaussians (HoG) or ResNet embeddings} $f(X)$ which is defined as follows:

\begin{equation}
	Cos_{Sim} = Cos_{Sim}(HoG(A), HoG(B))
\end{equation}

If the embedding of the ResNet function $f(X)$ is employed, the cosine similarity $Cos_{Sim}$ is computed as follows:	
\begin{equation}
	Cos_{Sim} = f(HoG(A), HoG(B))
\end{equation}

The Spearman correlation is defined as follows:
\begin{equation}
	\rho = Spearman(Cos_{Sim}, MCC)  
\end{equation}

\begin{table}[h]
	\centering
	\resizebox{\columnwidth}{!}{%
		\begin{tabular}{c c c c}
			\hline
			\textbf{ID} & \textbf{Dataset Name} & \textbf{Material} & \textbf{Publisher} \\
			\hline
			1 & CF\_t\_8 & Textile (CF) & Fraunhofer \\
			2 & CF\_80\_bright & Textile (CF) & Fraunhofer \\
			3 & CF\_80\_dark\_1 & Textile (CF) & Fraunhofer \\
			4 & CF\_80\_dark\_2 & Textile (CF) & Fraunhofer \\
			5 & CF\_80\_dark\_3 & Textile (CF) & Fraunhofer  \\
			6 & CF\_80\_dark\_4 & Textile (CF) & Fraunhofer \\
			7 & CF\_80\_dark\_5 & Textile (CF) & Fraunhofer \\
			8 & CF\_RefSet & Textile (CF) & Fraunhofer \\
			9 & CF\_RefSet\_Sm\_Dark & Textile (CF) & Fraunhofer \\
			10 & CF\_RefSet\_Sm\_Light & Textile (CF) & Fraunhofer \\
			11 & FabricDefectsAITEX & Textile & AITEX \\
			12 & KolektorSDD (kos10) & Electronics & Kolektor \\
			13 & MT\_Blowhole\_train & Metal & CN / Columbia \\
			14 & CF\_Sp0-0315 & Textile (CF) & Fraunhofer \\
			15 & CF\_Sp0-0315Thrd & Textile (CF) & Fraunhofer \\
			16 & CF\_Sp0-0315Thrd256 & Textile (CF) & Fraunhofer \\
            17 & CF\_Sp2-0816 & Textile (CF) & Fraunhofer \\
			18 & Bottle\_Brkn\_Lg & Glas & MVTec \\
			19 & Bottle\_Brkn\_Sm & Glas & MVTec \\
			20 & Cable\_Missing & Electronics & MVTec \\
			21 & Capsule & Plastic & MVTec \\
			22 & Carpet & Textile & MVTec \\
			23 & Grid\_Thread & Metal & MVTec \\
			24 & Hazelnut\_Crack & Food & MVTec \\
			25 & Metal\_Nut & Metal & MVTec \\
			26 & Pill\_Crack & Plastic & MVTec \\
			27 & Screw\_Scratch & Metal & MVTec \\
			28 & Tile\_Crack & Tile & MVTec \\
			29 & Toothbrush\_Sm & Plastic & MVTec \\
			30 & Wood\_Scratch & Wood & MVTec \\
			31 & Zipper\_Rough & Clothing & MVTec \\
			32 & Leather & Clothing & MVTec \\
			33 & Pul\_Resin & Fluid Resin & Fraunhofer \\
			34 & Pul\_Resin\_Augtd & Fluid Resin & Fraunhofer \\
			35 & Pul\_Window & Fluid Resin & Fraunhofer \\
			36 & severstal-steel & Metal & Severstal \\
			37 & Transistor & Electronics & MVTec \\
			38 & RoadCracks\_Small & Roads & CN Academy \\
			\hline
		\end{tabular}%
	}
	\caption{Numbered list of all industrial datasets}
	\label{tab:datasets}
\end{table}

Tab. ~\ref{tab:datasets} lists all datasets used in this study with their ID and dataset name mapping.

The \textit{Matthews Correlation Coefficient (MCC)} is a balanced metric for binary classification that accounts for all entries of the confusion matrix in the range $[-1;1]$:

\[
\mathrm{MCC} = \frac{TP \cdot TN - FP \cdot FN}{\sqrt{(TP + FP)(TP + FN)(TN + FP)(TN + FN)}}
\]

\textit{Comments on Dataset Quality:}

For this study, the resulting graph representations composed of image filters were converted into \texttt{hdev} code and subsequently applied across different datasets as mentioned in the experiments section of the paper. In a few rare cases, the evaluation metric describing the intersection between the filter predictions and the ground truth labels deviated slightly from the original results returned by the CGP image filter framework which this study is based on.

These discrepancies may arise from minor differences in numerical computations between the C\# implementation of the CGP filter framework, the \texttt{hdev} conversion (HDev code), and our experimental framework implemented in Python. To ensure transparency and reproducibility, we provide the filter pipelines for each dataset as references throughout all experiments reported in this paper.

The small deviations from the original MCC values reported in previous work do not affect the validity of the statistical tests conducted in this study, as all analyses are performed consistently within the datasets, filter pipelines, and code base specifically created for this experimentation.

\section*{Image Complexity Metrics}\label{appx:image-complexity}
This appendix section gives an overview on the mentioned image complexity metrics and their mathematical definitions.

\paragraph{Edge Density (ED)}
Edge density can be defined as the ratio of the total length of detected edges to the total area of the image. We denote the length and width of a set of edges $E$ detected after applying the canny edge operator as $L$ and $W$, and the area of the image as $S(I) = H(I) \times W(I)$. The mathematical definition of edge density (ED) can be expressed as:
\[
ED = |E| / S(I)
\]

where $ED$ represents the edge density of the image. The resulting binary edge image is summed to obtain the total number of edge pixels. The edge density $ED$ is computed by dividing the sum of edge pixels by the total number of pixels in the image height $x$ width).

\paragraph{Histogram Entropy (HE)}
Let $H(X)$ be the histogram of a discrete random variable $X$ with $n$ possible values $x_1, x_2, \dots, x_n$. The probability mass function of $X$ is given by $p(x) = H(x)/N$, where $H(x)$ is the frequency of occurrence of the value $x$ in the histogram and $N$ is the total number of samples. The entropy of the histogram, denoted as $H(X)$, is computed as:
\[
H(X) = -\sum_{i=1}^{n} p(x_{i}) log_{2}{p(x_{i})}
\]
The entropy value quantifies the amount of information or randomness in the distribution.

\paragraph{Texture Features (TF)}
Texture complexity is quantified using the \emph{contrast} property derived from a Grey Level Co-occurrence Matrix (GLCM). The GLCM is computed using the \texttt{scikit-image} implementation.\footnote{See scikit-image documentation: \url{https://scikit-image.org/docs/stable/auto_examples/features_detection/plot_glcm.html}}

The contrast measure captures the amount of local gray-level variation in an image and is defined as

\[
\mathrm{Contrast}
=
\sum_{i,j=0}^{levels-1}
P_{i,j}(i-j)^2,
\]
where $P_{i,j}$ denotes the normalized co-occurrence probability of gray levels $i$ and $j$. Higher contrast values indicate stronger intensity differences between neighboring pixels and therefore a more complex texture structure.

Prior to the computation of the contrast measure, each GLCM is normalized such that the sum of all matrix entries equals one.

\paragraph{Frequency (FR)}
Let $H(X)$ be the histogram of a discrete random variable $X$ with $n$ possible values $x_1, x_2, ..., x_n$. The probability mass function of $X$ is given by $p(x) = H(x)/N$, where $H(x)$ is the frequency of occurrence of the value $x$ in the histogram and $N$ is the total number of samples.
The entropy of the histogram, denoted as $H(X)$, is computed as:
\[
H(X) = -\sum_{i=1}^{n} p(x_{i}) log_{2}{p(x_{i})}
\]
The entropy value quantifies the amount of information or randomness in the distribution.

\paragraph{Number of Superpixels (SP)} 
For SP, we denote an input image as $I$, and let $S = {s_1, s_2, .\dots s_N}$ represent the set of superpixels obtained from image $I$. Each superpixel $s_i$ consists of a collection of pixels within the image. To compute the number of superpixels we minimize the following energy function:
\[
E(S) = E_{data}(S) + \lambda * E_{smooth}(S)
\]

where $E\_{data}(S)$ quantifies the similarity of pixel values within each superpixel and the corresponding image region, and $E\_{smooth}(S)$ promotes spatial regularity and connectivity among neighboring superpixels. $\lambda$ is a trade-off parameter that balances the influence of data fidelity and smoothness terms.

The goal is to find the optimal set of superpixels S that minimizes the energy function $E(S)$ through an iterative process. The resulting set of superpixels $S$ provides a partitioning of the image into compact regions.
\end{document}